\documentclass{scrartcl}

\RequirePackage{etoolbox}
\csdef{input@path}%
{%
 {sty/},
 {img/},
}

\usepackage{amsmath,amssymb,amsfonts}
\usepackage{amsthm}
\usepackage{mathtools}

\usepackage{booktabs}
\usepackage[dvipsnames]{xcolor}
\usepackage{graphicx}
\usepackage{siunitx}
\usepackage{url}
\usepackage{algorithm}
\usepackage{algorithmicx}
\usepackage{csquotes}
\usepackage{multirow}
\usepackage[colorlinks=true, citecolor=black, linkcolor=cyan]{hyperref}
\usepackage{glossaries-extra}
\usepackage{colortbl}
\usepackage{tikz}
\usepackage[top=2.0cm, bottom=4.0cm, left=2.0cm, right=2.0cm]{geometry}
\makenoidxglossaries

\newabbreviation{ar}{AR}{adversarial regularization}
\newabbreviation{adam}{Adam}{Adaptive Moment Estimation optimizer}

\newabbreviation{bl-ift}{BL-IFT}{bilevel learning with implicit differentiation}
\newabbreviation{bl-jfb}{BL-JFB}{bilevel learning with Jacobian free backpropagation}
\newabbreviation{cg}{CG}{conjugate gradient method}
\newabbreviation{ct}{CT}{computed tomography}
\newabbreviation{cnn}{CNN}{convolutional neural network}
\newabbreviation{crr}{CRR}{convex ridge regularizer}

\newabbreviation{gmm}{GMM}{Gaussian mixture model}
\newabbreviation{pdhgm}{PDHG}{primal-dual hybrid gradient}

\newabbreviation{em}{EM}{Expectation-Maximization}

\newabbreviation{epll}{EPLL}{expected patch log-likelihood}

\newabbreviation{fbp}{FBP}{filtered backprojection}
\newabbreviation{foe}{FoE}{fields of experts}

\newabbreviation{icnn}{ICNN}{input-convex neural network}
\newabbreviation{idcnn}{IDCNN}{input difference-of-convex neural network}
\newabbreviation{dc}{DC}{difference-of-convex}
\newabbreviation{ift}{IFT}{implicit function theorem}

\newabbreviation{jfb}{JFB}{Jacobian-free backpropagation}

\newabbreviation{lpd}{LPD}{learned primal-dual}
\newabbreviation{lpn}{LPN}{learned proximal network}
\newabbreviation{lsr}{LSR}{least-squares residual}
\newabbreviation{lar}{LAR}{local adversarial regularization}

\newabbreviation{mmse}{MMSE}{minimum mean-squared-error}
\newabbreviation{mse}{MSE}{mean squared error}
\newabbreviation{map}{MAP}{maximum a-posteriori}
\newabbreviation{mri}{MRI}{magnetic resonance imaging}
\newabbreviation{maid}{MAID}{method of adaptive inexact descent}
\newabbreviation{minres}{MINRES}{minimal residual method}
\newabbreviation{nett}{NETT}{network Tikhonov}
\newabbreviation{nf}{NF}{normalizing flow}
\newabbreviation{nmapg}{nmAPG}{nonmonotonic Accelerated Proximal Gradient algorithm}

\newabbreviation{psnr}{PSNR}{peak signal-to-noise ratio}
\newabbreviation{pnp}{PnP}{plug-and-play}
\newabbreviation{patchnr}{PatchNR}{patch normalizing flow regularizer}
\newabbreviation{patchml}{PatchML}{patch-based maximum likelihood}
\newabbreviation{pm}{PM}{proximal matching}

\newabbreviation{sm}{SM}{score matching}

\newabbreviation{relu}{ReLU}{rectified linear unit}
\newabbreviation{tdv}{TDV}{total deep variation}
\newabbreviation{tv}{TV}{total variation}

\newabbreviation{wcrr}{WCRR}{weakly-convex ridge regularizer}

\usepackage[noend]{algpseudocode}
\usepackage{commands_sebastian}

\usepackage{enumitem}
\setlist[itemize]{
  topsep=2pt,      
  parsep=0.5pt,   
}

\newcommand\blfootnote[1]{%
  \begingroup
  \renewcommand\thefootnote{}\footnote{#1}%
  \addtocounter{footnote}{-1}%
  \endgroup
}

\title{Learning Regularization Functionals for Inverse Problems: A Comparative Study}

\author{
Johannes Hertrich\textsuperscript{a,b,*}\and 
Hok Shing Wong\textsuperscript{c,*}\and 
Alexander Denker\textsuperscript{d}\and 
Stanislas Ducotterd\textsuperscript{e}\and 
Zhenghan Fang\textsuperscript{f}\and
Markus Haltmeier\textsuperscript{g}\and 
\smash{\v Z}eljko Kereta\textsuperscript{d} \and 
Erich Kobler\textsuperscript{h}\and
Oscar Leong\textsuperscript{i}\and
Mohammad Sadegh Salehi\textsuperscript{j}\and
Carola-Bibiane Schönlieb\textsuperscript{k}\and
Johannes Schwab\textsuperscript{l}\and 
Zakhar Shumaylov\textsuperscript{k}\and
Jeremias Sulam\textsuperscript{f}\and 
German Sh\^ama Wache\textsuperscript{m}\and 
Martin Zach\textsuperscript{e}\and 
Yasi Zhang\textsuperscript{i}\and
Matthias J. Ehrhardt\textsuperscript{c,\ding{61}} \and 
Sebastian Neumayer\textsuperscript{m,\ding{61}}
}
\date{}
\begin{document}
\maketitle

\begin{abstract}\noindent
In recent years, a variety of learned regularization frameworks for solving inverse problems in imaging have emerged.
These offer flexible modeling together with mathematical insights.
The proposed methods differ in their architectural design and training strategies, making direct comparison challenging due to non-modular implementations.
We address this gap by collecting and unifying the available code into a common framework.
This unified view allows us to systematically compare the approaches and highlight their strengths and limitations, providing valuable insights into their future potential.
We also provide concise descriptions of each method, complemented by practical guidelines.
\end{abstract}
\blfootnote{\noindent
\phantom{\textsuperscript{a}}\textsuperscript{a}Universit\'e Paris Dauphine-PSL, FR
\hspace{1em}
\textsuperscript{b}Inria Paris, FR
\hspace{1em}
\textsuperscript{c}University of Bath, UK
\hspace{1em}
\textsuperscript{d}University College London, UK
\hspace{1em}
\textsuperscript{e}École Polytechnique Fédérale de Lausanne, CH
\hspace{1em}
\textsuperscript{f}Johns Hopkins University, Baltimore, US
\hspace{1em}
\textsuperscript{g}University of Innsbruck, AT
\hspace{1em}
\textsuperscript{h}Johannes Kepler University Linz, AT
\hspace{1em}
\textsuperscript{i}University of California, Los Angeles, US
\hspace{1em}
\textsuperscript{j}Independent Scholar, UK
\hspace{1em}
\textsuperscript{k}University of Cambridge, UK
\hspace{1em}
\textsuperscript{l}University of Applied Sciences Kufstein, AT
\hspace{1em}
\textsuperscript{m}Chemnitz University of Technology, DE

\noindent
\textsuperscript{*}joint first author
\hspace{1em}
\textsuperscript{\ding{61}}joint last author
}

\section{Introduction}\label{sec:introduction}

Inverse problems are ubiquitous in imaging sciences.
As an example, \gls{mri} and X-ray \gls{ct} play a central role in many modern applications. 
Mathematically, the reconstruction is commonly modeled as a linear inverse problem~\cite{ribes2008linear}.
More precisely, we want to reconstruct an (unknown) image $\image \in \R^\dimimage$ from an observation $\data \in \R^\dimdata$ determined by the linear relation
\begin{equation}\label{eq:InvProb}
    \data = \forwardop\image + \vec n,
\end{equation}
where $\forwardop \in \R^{\dimdata \times \dimimage}$ encodes the underlying data acquisition process and the noise $\vec n\in\R^\dimdata$ accounts for imperfections in this description.
As $\forwardop$ is often ill-conditioned or non-invertible, the inverse problem \eqref{eq:InvProb} is ill-posed in the sense of Hadamard \cite{Hadamard1923} and reconstructing $\image$ from $\data$ is challenging.

A classical method to address ill-posedness is variational regularization, for which the unknown $\image$ is approximated by
\begin{equation}\label{eq:VarProb}
    \hat{\image}(\data) = \argmin \limits_{\image} \bigl\{ \datafit (\forwardop\image, \data) + \regparam \regularizer(\image) \bigr\}.
\end{equation}
In \eqref{eq:VarProb}, the data fidelity $\datafit \colon \R^\dimdata \times \R^\dimdata \to \R$ ensures data consistency, the regularizer $\regularizer \colon \R^\dimimage\rightarrow \R$ promotes desired properties of $\image$, and the regularization parameter $\regparam > 0$ balances the two.
There is a vast zoo of regularizers $\regularizer$ in the literature \cite{scherzer2009variational}.
A prominent example is the (anisotropic) \gls{tv} \cite{rudin1992nonlinear} $\regularizer(\image) = \|\grad \image\|_1$, which measures the $\ell_1$-norm of the discretized gradient.
The variational approach \eqref{eq:VarProb} leads to several desirable properties, e.g.,
\begin{itemize}
    \item \textbf{universality:} different forward and noise models can be incorporated;
    \item \textbf{data consistency:} the reconstruction $\hat{\image}(\data)$ satisfies \eqref{eq:InvProb} approximately, with control provided by the regularization parameter $\regparam$;
    \item \textbf{stability:} the data-to-reconstruction map $\data \mapsto \hat{\image}(\data)$ is often continuous.
    Namely, the noise $\vec n$ is not arbitrarily amplified in the reconstruction;
    \item \textbf{interpretability:} the regularization functional $R$ can be analyzed.
\end{itemize}
The literature on mathematical analysis for variational regularization methods is vast, see \cite{Benning2018actanumerica,ito2014inverse, scherzer2009variational} and the references therein.

In many situations, the variational approach \eqref{eq:VarProb} has a Bayesian interpretation.
There, the solution to the inverse problem \eqref{eq:InvProb} is formally defined as the posterior distribution of possible reconstructions $\image$ given some measurement \( \data \).
To this end, the image \( \image \in \R^\dimimage \) is modeled as a realization of a random variable \( X \sim \mathbb{P}_X \).
Reconstruction of \( \image \) from \( \data \) is then addressed by analyzing the posterior \( \mathbb{P}_{X \mid Y} \), which can be expressed via Bayes' theorem as
\begin{equation}\label{eq:BayesTheorem}
    \mathbb{P}_{X \mid Y} (\image \mid \data) \propto \mathbb{P}_{Y \mid X}(\data \mid \image) \mathbb{P}_{X}(\image).
\end{equation}
The conditional distribution \( \mathbb{P}_{Y \mid X} \) is usually available as it is induced by $\forwardop$ and the noise distribution.
Consequently, the challenge lies in finding accurate models of the prior \( \mathbb{P}_X \).
In our finite-dimensional setting, it is natural to assume that the distributions \(  \mathbb{P}_{X \mid Y}\), \( \mathbb{P}_{Y \mid X} \), and \( \mathbb{P}_{X} \) admit densities with respect to the Lebesgue measure, which we denote by \( p_{X \mid Y}\), \( p_{Y \mid X} \), and \( p_{X} \). 
There exist various statistical estimators of the posterior $p_{X \mid Y}$.
Among them, the \gls{map} estimator of $X$ given $Y=\data$, defined as
\begin{equation}
    \argmax_{\image \in \R^\dimimage}  p_{X\mid Y}(\image \mid \data) = \argmin_{\image \in \R^\dimimage} \bigl\{ -\log p_{Y \mid X}(\data \mid \image) - \log p_{X}(\image) \bigr\},
\end{equation}
recovers the variational problem \eqref{eq:VarProb} with \(p_{Y \mid X}(\data \mid \image) \propto \exp(-\datafit (\forwardop\image, \data)) \) and \(p_{X}(\image) \propto \exp(-\regparam \regularizer(\image)) \).
A second popular choice is the \gls{mmse} estimator, which can be shown to be the \emph{expectation} of the posterior \( p_{X \mid Y}\) rather than its maximum.

Over the past years, deep-learning-based approaches have become the state-of-the-art for solving inverse problems and there are many excellent reviews \cite{AMOS2019, habring2024neural, ongie2020deep, wang2020deep}.
Although they achieve impressive results, several concerns regarding their trustworthiness remain.
Recent works reveal troublesome issues that may arise for deep-learning-based approaches if the aforementioned \emph{desirable properties} are not met~\cite{antun2020instabilities,gottschling2020troublesome}.
In contrast, hand-crafted regularizers $\regularizer$ such as \gls{tv} are  theoretically founded but cannot achieve the same reconstruction quality as data-driven approaches.
We focus on the blend of these approaches, namely the learning of $\regularizer$ from data.
Occasionally, we write $\learnedreg$ to emphasize the dependence on the parameters $\parameters$. 
Below, we give a brief overview of the state-of-the-art in the learning of regularizers.
In Sections~\ref{sec:architectures} and \ref{sec:training}, we go into more detail for the regularizers and training methods contained in this comparison.

A pioneering learnable regularizer $\regularizer$ is the \gls{foe} \cite{RotBla2009}, which is the sum of 1D potentials composed with convolutional filters.
Recently, it was proposed to learn the \gls{foe} using linear splines, leading to the \gls{crr} \cite{GouNeuBoh2022} and \gls{wcrr} \cite{GouNeuUns2023}.
Another convex architecture is the \gls{icnn} \cite{AMOS2019} and its descendant, the input weakly-convex neural network \cite{shumaylov2024weakly}.
Following the idea of structured nonconvexity, these were extended to \glspl{idcnn} \cite{zhang2025learning}.
Examples of more complex multi-scale \gls{cnn} regularizers are the \gls{tdv}~\cite{KobEff2020}, the \gls{lsr} \cite{ZouLiuWoh2023}, and energy-based generative priors~\cite{ZaKo21CTEnergy}.
An alternative with the emphasis on sparse representations is dictionary learning \cite{bruckstein2009sparse,mairal2008supervised,tovsic2011dictionary}.
Such models are generalized to neural networks using a nonlinear representation via generative models \cite{alberti2024manifold,bora2017compressed,DufCamEhr2023,HabHol2022}.

A parallel development aims to instead learn the proximal operator
\begin{equation}
    \label{eq:proximals}
    \prox_\regularizer(\image) = \argmin_{\vec z} \bigl\{\tfrac 12 \|\image - \vec{z}\|^2_2 + \regularizer(\vec z)\bigr\},
\end{equation}
of $\regularizer$, which is central to proximal algorithms for solving \eqref{eq:VarProb}.
The interpretation of $\prox_\regularizer$ as a variational denoiser has inspired the popular \gls{pnp} approaches \cite{fermanian2022learned,HNS2021,laumont2022bayesian,PRTW2021,reehorst2018regularization,venkatakrishnan2013plug,Drunet2022}, which replace the $\prox_\regularizer$ in proximal algorithms with a learned denoiser.
Under certain conditions on its architecture, an underlying $\regularizer$ exists \cite{fang2023s,HurLec2022}.
Since this $\regularizer$ is only given implicitly, we are limited to proximal algorithms for solving \eqref{eq:VarProb} and tracking the objective values is difficult.
As an example, we discuss \glspl{lpn} in Section~\ref{sec:LPNs}.

Given a parametric regularizer $\regularizer_\parameters$, we need to learn its parameters $\parameters$ from data.
Towards this goal, many paradigms have been introduced over the past decades.
One notable paradigm is bilevel learning, which adapts the $\parameters$ such that the reconstruction \eqref{eq:VarProb} minimizes some loss.
This idea started with learning only the regularization parameter $\alpha$ in \eqref{eq:VarProb} \cite{calatroni2017bilevel,de2013image,de2017bilevel,haber2009numerical,kunisch2013bilevel}, and has been gradually lifted to learning regularizers.
The required gradients of the reconstructions with respect to $\parameters$ can be computed via implicit differentiation \cite{JiYanLia2021,LiaXioFet2018,ZucSac2022}, leading to the \gls{bl-ift} approach.
In practice, the optimization problem in \eqref{eq:VarProb} is only solved up to a certain precision.
The \gls{maid} \cite{salehi2024adaptively} and related works \cite{pedregosa2016hyperparameter, salehi2025ssvm} explicitly capture this inaccuracy.
If we instead use backpropagation to compute the gradients, the resulting method is commonly known as unrolling~\cite{MonLiEld2021}.
As the memory requirements grow linearly with the number of iterations of the deployed optimization algorithm, this is impractical in our setting.
Instead, we can deploy \gls{bl-jfb} \cite{BolPauVai2023,FunHeaLi2022} as efficient intermediate regime.

A second paradigm is based on distinguishing desirable and undesirable images a priori, without actually solving \eqref{eq:VarProb}.
This is reminiscent of classification with two classes.
Prominent examples include contrastive divergence~\cite{ZaKo21CTEnergy}, \gls{ar}~\cite{lunz2018adversarial, MukDit2021} and \gls{nett}~\cite{LiSch2020}.
During training, these approaches are not linked to the variational problem \eqref{eq:VarProb}, and require the selection of a suitable regularization parameter $\alpha$ for the inverse problem at hand.

A third paradigm arises from the Bayesian viewpoint \eqref{eq:BayesTheorem} and the interpretation of \eqref{eq:VarProb} as the \gls{map} estimator.
Under this framework, learning $\regularizer$ amounts to estimating the prior $p_X$.
Several authors construct $\regularizer$ by leveraging generative models \cite{helminger2021generic,wei2022deep}.
To reduce the computational effort and required data, \gls{epll} \cite{zoran2011learning}, \gls{lar} \cite{prost2021learning} and \gls{patchnr} \cite{altekruger2023patchnr} propose to instead approximate a patch distribution, see \cite{piening2024learning} for an overview.
Alternatively, we can approximate $p_X$ by the density $p_\sigma \coloneq p_{X_\sigma}$ of $X_\sigma=X+\sigma\eta$ with $\eta\sim\mathcal N(0,I)$.
Then, $\regularizer$ can be learned with a denoising loss via Tweedie’s formula~\cite{Mi61,Ro56}, which links $p_{\sigma}$ with the \gls{mmse} estimator of $X$ given $X_\sigma$.
The resulting training method is called \gls{sm} and several variants have been proposed~\cite{Hy05, KaEl22denoising,  RoBl22,  Vi11, ZaKo23ProductGMDM}.
Tweedie's formula also induces the popular gradient-step denoiser \cite{cohen2021has,hurault2022gradient,romano2017little}.

\textbf{Outline}
In Section~\ref{sec:architectures}, we review existing regularization architectures.
In Section~\ref{sec:training}, we discuss the training methodologies, ranging from bilevel learning to contrastive learning, and distribution matching approaches.
Section~\ref{sec:setup-exp} describes the setup for our numerical comparison.
In particular, this includes the optimization algorithm, forward models, datasets, and evaluation metrics.
Section~\ref{sec:results} presents the numerical results, which are discussed and interpreted in Section~\ref{sec:discussion}.
There, we also outline limitations and possible extensions.
Finally, conclusion are drawn in Section~\ref{sec:conclusions}.

\section{Overview of Regularizer Architectures} \label{sec:architectures}
First, we review various regularization architectures, which are summarized in Table~\ref{tab:Regularizers}.
These architectures vary in terms of parameter count, complexity, and convexity properties. 
For algorithmic convenience, we focus on differentiable $\regularizer$, though all architectures can be used with non-smooth activations.
Unless stated otherwise, $\grad \regularizer(\image)$ is computed using automatic differentiation.

\begin{table}[tbp]
\centering
\caption{Regularizer architectures and their parameter count as implemented for this chapter.\label{tab:Regularizers}}
\setlength\tabcolsep{5pt}
\begin{tabular}{llllllll}
\toprule
  & Convex  & Parameters & Backbone & Reference & Description \\ 
\midrule
\gls{crr} & \cmarkc & $\approx$ 15k & \gls{cnn} & \cite{GouNeuBoh2022} & Section \ref{sec:FoE} \\
\gls{wcrr} & \xmarkc & $\approx$ 15k & \gls{cnn} & \cite{GouNeuUns2023} & Section \ref{sec:FoE} \\
\gls{icnn} & \cmarkc & $\approx$ 26k & \gls{cnn} & \cite{MukDit2021} & Section \ref{sec:ICNN} \\
\gls{idcnn} & \xmarkc & $\approx$ 53k 
& \gls{cnn} & \cite{zhang2025learning} & Section \ref{Sec:IDCNN} \\
\gls{epll} & \xmarkc & $\approx$ 280k & dictionary learning & \cite{zoran2011learning} & Section \ref{Sec:PatchBased} \\
\gls{patchnr} & \xmarkc & $\approx$ 3M & normalizing flow & \cite{altekruger2023patchnr} & Section \ref{Sec:PatchBased} \\
\gls{cnn} & \xmarkc & $\approx$ 200k & \gls{cnn} & \cite{prost2021learning} & Section \ref{Sec:PatchBased} \\
\gls{tdv} & \xmarkc & $\approx$ 400k & UNet & \cite{KobEff2020,KobEff2021} & Section \ref{sec:TDV} \\
\gls{lsr} & \xmarkc & $\approx$ 4M & DRUNet & \cite{ZouLiuWoh2023} & Section \ref{sec:LSR} \\
\gls{lpn} & \xmarkc & $\approx$ 4M & UNet & \cite{fang2023s} & Section \ref{sec:LPNs} \\
\bottomrule
\end{tabular}
\end{table}

\subsection{Fields-of-Experts Regularizer} \label{sec:FoE}
\begin{figure}[tb]
    \centering
    \includegraphics[width=\linewidth]{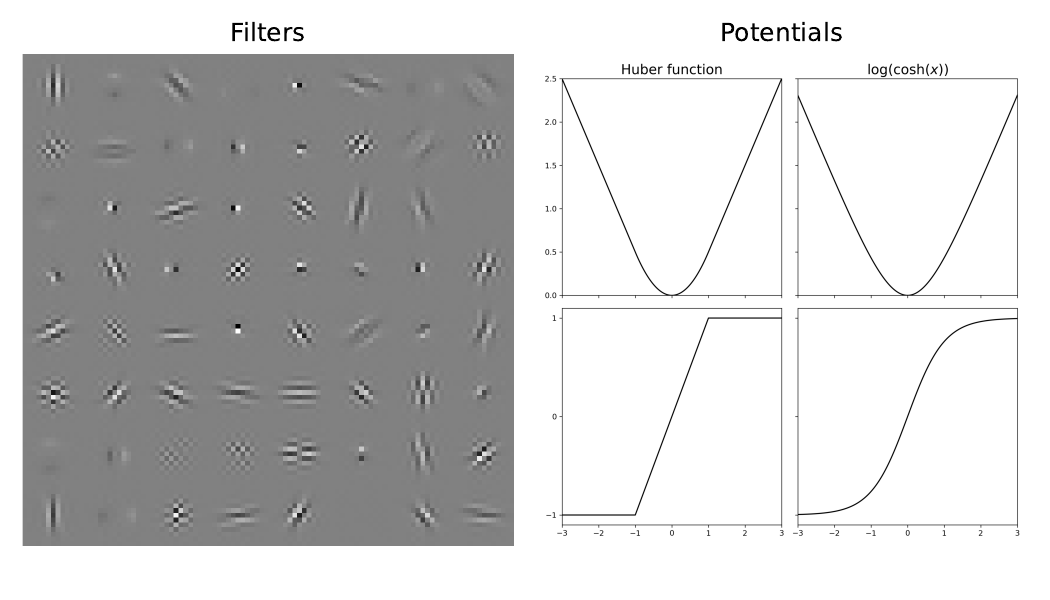}
    \caption{Left: The 64 filter impulse responses of a learned ridge regularizer. 
    Right: The two discussed potentials $\psi^1$ and their derivative $\varphi^1=(\psi^1)'$.}
    \label{fig:RR}
\end{figure}

In \cite{GouNeuBoh2022,GouNeuUns2023}, the authors discuss learning specific instances of the \glsxtrfull{foe} \cite{RotBla2009}, which takes the general form
\begin{equation}\label{eq:RegBase}
    \regularizer(\image) = \sum_{j=1}^{c} \langle \ones, \psi_j(\mat{W}_{j} \image)\rangle.
\end{equation}
For each of the $c$ filters, the potentials $\psi_j\colon \R \to \smash{\R^+}$ are applied componentwise and \smash{$\mat{W}_{j} \colon \R^\dimimage \to \R^\dimimage$} are convolutions.
Hence, $\regularizer$ is a spatial penalization of multiple filter responses.
Choosing $c = 2$, $\mat{W}_1 = \mat{D}_\mathrm{x}$, $\mat{W}_2 = \mat{D}_\mathrm{y}$ and $\psi_1= \psi_2 = \vert \cdot \vert$ leads to the \gls{tv} regularizer \cite{rudin1992nonlinear}, which often serves as a baseline.
If $\psi_j \in C^1(\R)$, then
\begin{equation}
\label{eq:gradmodel}
    \grad \regularizer(\image) = \sum_{j=1}^c \mat W_j^T \psi_j'(\mat W_j \image).
\end{equation}
In Figure \ref{fig:RR}, we visualize a specification of the \gls{foe} with learned $\psi_j$ and $\mat{W}_j$.

The authors of \cite{GouNeuBoh2022,GouNeuUns2023} parameterize the $(\mat W_j)_{j=1}^c$ as a multi-convolution (an instance of linear neural networks \cite{AroCohGol2019,BalHor1995}).
More precisely, to efficiently explore a large field of view, they decompose $(\mat W_j)_{j=1}^c$ into a composition of zero-padded convolutions with kernels of size $k \times k$ and with an increasing number of output channels.
The kernels of the first layer have zero mean.
Moreover, $\Vert (\mat W_j)_{j=1}^c \Vert_2= 1$ is required to avoid scaling ambiguities.
This constraint is implemented via spectral normalization based on power iterations.
An efficient estimation in terms of the discrete Fourier transform is given in \cite{GouNeuUns2023}. 

The authors of \cite{GouNeuUns2023} parameterize $\psi_j$ as ${\psi_j}(\scalarA) = \smash{1/\alpha_j^2 \psi^\beta(\alpha_j \scalarA)}$, where the learnable $\alpha_j \in \R$ adapt a shared potential $\psi^\beta$.
Here, the division by $\alpha_j^2$ ensures that the maximum of the (weak) derivative \smash{$\psi_j''$} is independent of $\alpha_j$.
Using a shared profile makes $\regularizer$ more interpretable and easier to analyze.
In particular, if $\psi^\beta$ is convex---which is the case when $(\psi^\beta)'' \geq 0$ a.e.---then $\regularizer$ is convex which in turn guarantees that the objective function in \eqref{eq:VarProb} is convex. 
In contrast to \cite{GouNeuBoh2022,GouNeuUns2023}, which use learnable splines~\cite{DGB2022} to parameterize $\psi^\beta$, we simply use the Huber function (Moreau envelope of the $\ell_1$ norm)
\begin{equation}\label{eq:MoreauEnvelope}
    \psi^\beta(\scalarA) = \begin{cases}
        \vert \scalarA \vert  - \frac{\beta^{-1}}{2}\quad & \vert \scalarA \vert > \beta^{-1},\\
        \frac{\beta}{2}\scalarA^2 \quad & \vert \scalarA \vert \leq \beta^{-1}
    \end{cases}
\end{equation}
with a learnable parameter $\beta$.
To ensure that $\regularizer \in C^\infty(\R^\dimimage)$, we can deploy $\psi^\beta(\scalarA) = \beta^{-1}\log(\cosh(\beta\scalarA))$ instead.
By computing the derivatives, we directly verify that both options lead to a convex $\regularizer$. In the denoising case where $\forwardop = \identity$, the objective in \eqref{eq:VarProb} remains convex even if $(\psi^\beta)'' \geq -1$ a.e., that is, if $\psi^\beta$ is $1$-weakly convex.
In this setting, we instead use $\tilde \psi^\beta(\scalarA) = \psi^\beta(\scalarA) - \psi^1(\scalarA)$ with $\beta\geq 0$ and one of the $\psi^\beta$ from above.
In accordance with \cite{GouNeuBoh2022,GouNeuUns2023}, we refer to these two instances of the \gls{foe} model as \gls{crr} and \gls{wcrr}, respectively.
The overall set of parameters $\parameters$ is given by the convolutions kernels $W_j$ and the parameters of the potentials $\psi_j$ for $j=1,\ldots,c$.

One possibility to make the architecture \eqref{eq:RegBase} more flexible is spatial adaptivity, namely to replace the constant vector $\ones$ by spatially varying weights $\boldsymbol \Lambda_j$.
These can be derived from the data $\data$ as done in \cite{NeuAlt2024,NeuPouGou2023,PouKobUns2025,PouNeuUns2024}.
Finally, a theoretical analysis for regularizers of the form \eqref{eq:RegBase} is performed in \cite{NeuAlt2024}.

\subsection{Convolutional Neural Network}\label{sec:CNN}
Often, the starting point for constructing a learnable regularizer $\regularizer \in C^1(\R^\dimimage)$ is a \gls{cnn}.
Here, $\regularizer=\vec z_\nlayers$, where $\vec z_\nlayers$ is defined recursively via
\begin{equation}\label{eq:feedforward}
    \vec z_1 = \psi_0(\mat{V}_0\image + \vec b_0),\quad \vec z_{i+1} = \psi_i(\mat{W}_i\vec z_i+\mat{V}_i \image +\vec b_i), \quad i=1,\dots,\nlayers-1,
\end{equation}
with parameters
$\parameters=\{(\mat{W}_i)_{i=1}^{\nlayers-1},(\mat{V}_i)_{i=0}^{\nlayers-1},(\vec b_i)_{i=0}^{\nlayers-1}\}$ and componentwise non-linear activation functions $\psi_i \in C^1(\R)$.
The operators $\vec W_i$, $\vec V_i$ denote standard (linear) transformations  such as convolutional or (averaged ) pooling layers.
Since $\vec z_\nlayers$ must be scalar, the output dimension of $\mat{W}_{\nlayers-1}$ and $\mat{V}_{\nlayers-1}$ must be one.
Typical examples for $\psi_i$ include
\begin{itemize}
    \item the softplus $\scalarA \mapsto \log(1 + \exp(\scalarA))$;
    \item the hyperbolic tangent $
    \scalarA \mapsto \tanh(\scalarA) = \frac{\exp(\scalarA) - \exp(-\scalarA)}{\exp(\scalarA) + \exp(-\scalarA)}$;
    \item the sigmoid $\scalarA \mapsto 1/(1+\exp(-\scalarA))$;
    \item the non-differentiable \gls{relu} $\scalarA \mapsto \max(\scalarA, 0)$ can be approximated by the smoothed (Moreau envelope) surrogate
    \begin{equation}\label{eq:huber_relu}
    \psi^\beta(\scalarA)=\begin{cases}
        0 &\text{ if }\scalarA\leq0,\\
        \frac{\beta}{2}\scalarA^2 &\text{ if }0<\scalarA<\beta^{-1},\\
        \scalarA-\frac{\beta^{-1}}{2} &\text{ otherwise}.
    \end{cases}
    \end{equation}
\end{itemize}
The \gls{foe} \eqref{eq:RegBase} is an instance of \eqref{eq:feedforward} with one hidden layer.
In the next sections, we discuss more complex choices.
The multi-scale architectures that we discuss in Sections \ref{sec:TDV} and \ref{sec:LSR} also include skip connections into \eqref{eq:feedforward}.

\subsection{Input Convex Neural Network}\label{sec:ICNN}
Several approaches to learning both convex as well as nonconvex $\regularizer$ are based on \glsxtrfullpl{icnn} \cite{amos2017icnn}.
There, we constrain the $\vec W_i$ and $\psi_i$ such that \eqref{eq:feedforward} is a convex functional.
For this, we use the fact that convexity is preserved under non-negative linear combinations and composition with a convex non-decreasing function.
Thus, \eqref{eq:feedforward} is convex if the $\psi_i$ are convex and non-decreasing, and the $\vec W_i$ have non-negative entries.
The activation functions $\psi$ from Section \ref{sec:CNN} satisfy these properties.
In \cite{amos2017icnn}, the non-negativity of the $\vec W_i$ is enforced via zero clipping, i.e., by projecting the entries to the non-negative numbers after every training step.
An alternative is to use positive parameterizations of the weights, for example via the exponential function.

For our experiments, we use an \gls{icnn} with two layers and no skip connections, namely
\begin{equation}\label{eq:ICNN}
\regularizer(\image)=\sum_{j=1}^ca_j\langle\ones, \psi_j(\vec W_{2, j} \vec z_j)\rangle, \text{ with } \vec z_j=\psi_j(\vec W_{1, j} \image).
\end{equation}
In this model, the $\vec W_{i, j}$ as well as the coefficients $a_j$ are learnable.
The $\psi_j$ are chosen as the smoothed \gls{relu} with learnable $\beta$.
More layers were difficult to train and did not lead to significant improvements in our experiments.

The idea of using \glspl{icnn} as regularizers goes back to \cite{MukDit2021}.
Beyond this, \Glspl{icnn} can be used to model $\regularizer$ as the difference of convex functions~\cite{cohen2021has, zhang2025learning}, see Section~\ref{Sec:IDCNN}.
Moreover, they enable strategies for learning a proximal operator, see Section~\ref{sec:LPNs}.
Another extension are weakly-convex \glspl{icnn} \cite{shumaylov2024weakly}.

\subsection{Input Difference-of-Convex Neural Network} \label{Sec:IDCNN}

While convexity of $\regularizer$ leads to a convex objective in \eqref{eq:VarProb}, it limits the expressiveness of $\regularizer$.
One possibility for architectures with structured nonconvexity and thus more modeling flexibility is the \gls{dc} functions framework \cite{cohen2021has, zhang2025learning}.
There, $\regularizer$ is written as
\begin{align}
    \regularizer(\image) = \regularizer_1(\image) - \regularizer_2(\image),
\end{align}
where $\regularizer_1$ and $\regularizer_2$ are convex.
In this case, we say that $\regularizer$ is a \gls{dc} function.
Several popular nonconvex, hand-crafted sparsity penalties fall in this class, including SCAD~\cite{fan2001variable}, MCP \cite{zhang2010nearly}, the logarithmic penalty \cite{mazumder2011sparsenet}, and the difference between the $\ell_1$-norm and $\ell_2$-norm \cite{yin2015minimization}.
The class of \gls{dc} functions is broad and includes special classes of nonconvex functions, such as weakly-convex functions.
Moreover, this class is closed under natural operations, such as linear combinations, multiplication, and division~\cite{le2018dc}.
To learn a \gls{dc} regularizer $\regularizer$, one can take $\regularizer$ as the difference of two \glspl{icnn} $\regularizer_1$ and $\regularizer_2$, see Section~\ref{sec:ICNN}. We refer to this as an \glsxtrfull{idcnn}.

Due to the \gls{dc} structure, we could leverage specialized algorithms when solving the corresponding variational problem \eqref{eq:VarProb}.
In particular, when the data fidelity $\image \mapsto \datafit(\forwardop\image, \data)$ is convex, we can write the objective as the difference of $F_1(\image) \coloneq \datafit(\forwardop\image, \data) + \regparam \regularizer_1(\image)$ and $F_2(\image) \coloneq \regparam \regularizer_2(\image)$.
Then, we can use the \gls{dc} algorithm \cite{le2018dc}.
At each step, this algorithm linearizes the concave part $-F_2(\image)$ and minimizes the resulting convex majorization of $F_1 - F_2$.

\subsection{Patch-Based Architectures} \label{Sec:PatchBased}
Many priors are constructed such that they only use local information in images.
Patch-based methods \cite{piening2024learning} exploit this idea by splitting the input into small regions of size $l\times l$, which we call patches.
Well-known denoising algorithms based on this principle are non-local means~\cite{buades2005review} and BM3D~\cite{dabov2007image}. 
Here, we consider patch-based regularizers $\regularizer$, namely the \glsxtrfull{epll} \cite{zoran2011learning}, \glsxtrfull{patchnr} \cite{altekruger2023patchnr}, and \glsxtrfull{lar} \cite{prost2021learning}. 
A similar approach based on diffusion models was proposed in~\cite{hu2024learning}.
Formally, we define the patch extractor $E_i\colon\R^d\to\R^k$ with $k=l^2$ and $i=1,\dots,s$, which extracts the $i$-th patch from the input image.
Then, we define the regularizer $\regularizer_\parameters$ as
\begin{align}\label{eq:PatchReg}
    \regularizer_\parameters(\image) = \frac{1}{\npatches} \sum_{i=1}^\npatches r_\parameters(E_i\image), 
\end{align}
where $r_\parameters\colon\R^k\to \R$ is a (learnable) regularizer on patches.
Since every image contains a large number of patches, $r_\parameters$ can be trained on very small datasets.

Both \gls{epll} and \gls{patchnr} rely on statistical models to design $r_\parameters$. 
More precisely, we define $r_\parameters(x) \coloneq -\log p_{\parameters}(x)$, where $p_{\parameters}$ is a  distribution on the space of patches.
The latter is usually learned using \gls{patchml} estimation.
\gls{epll} typically models $p_{\parameters}$ as a \gls{gmm} with $c$ components, giving 
\begin{align}
    \log p_{\parameters}(x) = \log\biggl( \sum_{i=1}^c a_i g(x; \mu_i, \Sigma_i) \biggr),
\end{align}
where $g(x; \mu_i, \Sigma_i)$ is the multivariate Gaussian density with weights, means, and covariances $\parameters=\{(a_i,\mu_i,\Sigma_i)\}_{i=1}^c$.

Instead, $p_{\parameters}$ can be chosen as, e.g., constrained or generalized \glspl{gmm}~\cite{deledalle2018image, houdard2018high,NHAB2023}, distributions incorporating multi-scale \cite{papyan2015multi} or sparsity elements \cite{sulam2015expected}.
Traditionally, the variational problem \eqref{eq:VarProb} is solved via half-quadratic splitting, which alternates between updates over the patches and the entire image.
This involves several approximations, which introduces additional regularization.

\Gls{patchnr} leverages \glspl{nf} to parameterize $p_{\parameters}$.
More precisely, given some latent distribution $\mathbb{P}_Z$ (typically $z \sim \mathcal{N}(0, \identity)$) and a diffeomorphism $T_{\parameters}\colon \R^k \to \R^k$, we define $\mathbb P_{\parameters}$ via the push forward operator as $\mathbb{P}_{\parameters} = (T_{\parameters})_\# \mathbb{P}_Z$.
Its density can be calculated as 
\begin{align}
    \label{eq:change_of_var}
    p_{\parameters}(x) = p_z(T_{\parameters}^{-1}(x)) | \det J_{T_{\parameters}^{-1}}(x) |,
\end{align}
where \smash{$J_{T_{\parameters}^{-1}}$} denotes the Jacobian of $T_{\parameters}^{-1}$.
Evaluating $p_{\parameters}$ requires an efficient inverse $T_{\parameters}^{-1}$ and a tractable Jacobian determinant. 
To this end, $T_{\parameters}$ is commonly implemented as invertible neural network with affine coupling layers \cite{dinh2017density}, where
the input $\vec x \in \R^k$ is split into two parts as $\vec x =[\vec x_1, \vec x_2] \in \R^{k_1+k_2}$ and $T_\parameters$ and $T_\parameters^{-1}$ are defined as
\begin{equation}\label{eq:INNPass}
	\begin{array}{ll}
    \text{\textit{{Forward Pass}}}\\
	\vec z_1 = \vec x_1 \\
	\vec z_2 = \vec x_2 \odot \exp(s(\vec x_1)) + t(\vec x_1) \\
	\end{array} 
    \Leftrightarrow \begin{array}{ll}
        \text{\textit{{Inverse Pass}}}\\
	\vec x_1 = \vec z_1 \\ 
        \vec x_2 =  (\vec z_2 - t(\vec z_1)) \odot \exp(-s(\vec z_1)),
	\end{array}
\end{equation}
where $s\colon \R^{k_1} \to \R^{k_2}$ and $t\colon \R^{k_1} \to \R^{k_2}$ are arbitrary (unconstrained) neural networks.
Coupling layers are typically stacked alternatingly, i.e., components left unchanged in one layer are updated in the next.

Finally, we can define $r_\parameters$ in \eqref{eq:PatchReg} by a padding-free \gls{cnn}, see \cite{prost2021learning}. Due to the valid padding the \gls{cnn} takes an input patch of size $l\times l$ and returns a single number as an output. Consequently, the patch size $l$ corresponds to the receptive field of the \gls{cnn} determined by the kernel size and the number of layers. If we apply the same \gls{cnn} to a larger image, the output corresponds to a matrix with the output of $r_\parameters$ for all patches in the image. Hence, computing $\regularizer_\parameters$ can be evaluated by applying the \gls{cnn} and averaging over the outputs. 
The authors of \cite{prost2021learning} propose to learn such \glspl{cnn} with \gls{lar} training, but most of the other training methods of Section~\ref{sec:training} can be applied as well.

Many patch-based methods (including \gls{epll}, \gls{patchnr} and padding-free \glspl{cnn}) are prone to boundary artifacts since pixels at  the image boundary are covered from fewer patches than interior pixels.
As a remedy, we pad the input image in \eqref{eq:PatchReg} by $l-1$ pixels, where $l$ is the patch size.

\subsection{Total Deep Variation} \label{sec:TDV}

In analogy to the \gls{foe}~\eqref{eq:RegBase}, \glsxtrfull{tdv}~\cite{KobEff2020} extends the model by incorporating a non-linear, multi-scale feature transform based on a \gls{cnn}.
For $\image\in\R^\dimimage$, \gls{tdv} is defined as
\begin{equation}\label{eq:RegTDV}
    \regularizer_\mathrm{\gls{tdv}}(\image) = \langle \ones, \psi(\mat w \Psi(\mat{W} \image))\rangle,
\end{equation}
where $\mat{W}\in\R^{\dimimage c\times \dimimage}$ is composed of $c$ convolutions, $\Psi \colon \R^{\dimimage c}\to\R^{\dimimage c}$ is a multi-scale \gls{cnn}, $\mat w\in\R^{d\times dc}$ is a $1\times1$ convolution, and $\psi\colon \R\to\R$ is a component-wise potential such as $\psi(x) = \frac12 x^2$ or $\psi(\scalarA) = \log (\cosh (\scalarA))$.
Compared to other deep network-based regularizers, see for example \cite{cohen2021has,ZouLiuWoh2023}, the architecture \eqref{eq:RegTDV} increases and reduces the channel number through $\mat W$ and $\mat w$, respectively.

\begin{figure}%
\centering
\includegraphics[width=.6\linewidth]{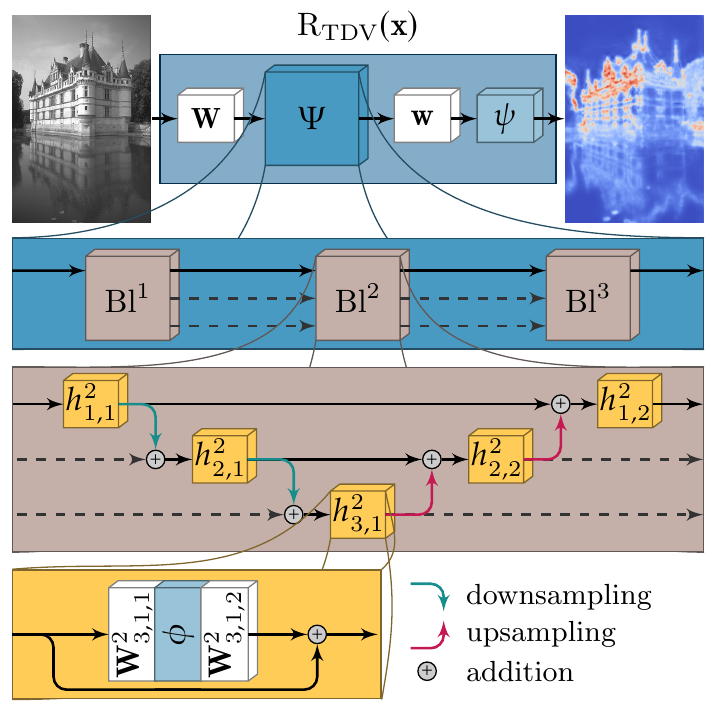}
\caption[Visualization of the \gls{tdv} regularizer]{Visualization of \gls{tdv}$_3^3$. 
On the highest level, an energy value is assigned to every pixel by applying a \gls{cnn}.
The \gls{cnn}~$\Psi$ (blue) is composed of three UNet-like macro-blocks (gray).
Each macro-blocks consists of five micro-blocks (yellow) with residual connections.}
\label{fig:regTDV}
\end{figure}

The computational structure of $\Psi$ follows a hierarchical design and is visualized in Figure~\ref{fig:regTDV}.
Specifically, $\Psi$ is composed of $b$ sequential  UNet type~\cite{ronneberger2015u} macro-blocks~$\mathrm{Bl}^i, i\in\{1,\ldots, b\}$ with $a$ scales.
We denote this as \gls{tdv}$^b_a$.
On each scale of  $\mathrm{Bl}^i$, we apply the residual micro-blocks
\begin{equation}
    h_{j,k}^i(\image) = \image + \mat{W}_{j,k,2}^i \phi(\mat{W}_{j,k,1}^i \image), \qquad j\in\{1,\ldots,a\},\, k\in\{1,2\}, 
\end{equation}
where $\mat{W}_{j,k,1}^i,\mat{W}_{j,k,2}^i\in\R^{dc\times dc}$ are convolutions and $\phi\colon\R\to\R$ is a component-wise activation function.

While \cite{KobEff2020,KobEff2021} originally used Student-t type activation functions, we apply the softplus function for more stable training dynamics.
The downsampling and upsampling operations within $\Psi$ are implemented by strided $3\times3$ convolutions and transposed convolutions, respectively, combined with an anti-aliasing blur kernel, following~\cite{Zh19}.
Instead of symmetric boundary handling as in \cite{KobEff2020}, we use zero boundary without a performance decrease.

Compared to the \gls{foe}~\eqref{eq:RegBase}, the \gls{tdv}~\eqref{eq:RegTDV} uses a \emph{non-linear} feature transform~$\Psi$ prior to the application of the potential $\psi$.
Its hierarchical multi-scale architecture enables the extraction of more complex features over large spatial neighborhoods.
Based on a mean-field control interpretation,  robustness and stability results with respect to perturbations in both measurements and parameters were derived \cite{KobEff2021}.
Successful applications of the \gls{tdv} include accelerated \gls{mri} \cite{NarEff21}, structured illumination microscopy~\cite{WanFan23}, exit wave reconstruction in transmission electron microscopy~\cite{PinKob21}, and learning of binary sampling patterns for single-pixel imaging~\cite{TudDen25}.
In combination with a patch-wise Wasserstein distance, \gls{tdv} can also be used in unsupervised settings \cite{PinKob21b}.

\subsection{Least Squares Residual Regularizer}\label{sec:LSR}

A common principle is that reconstructions live in a set (or manifold) $\set M$, which can be characterized as the fixed points of a mapping $U \colon \R^\dimimage \to \R^\dimimage$.
Then, penalizing the residuals $\image - U(\image)$ yields a regularizer
\begin{equation}\label{eq:LSR}
    \regularizer(\image) = \frac12 \Vert \image - U(\image) \Vert^2, 
\end{equation}
which we refer to as the \glsxtrfull{lsr} regularizer.
The minimum of \eqref{eq:LSR} is attained when $U(\image)= \image$, that is, when $\image$ is a fixed point.
Further, its gradient is given by 
\begin{equation}
    \grad \regularizer(\image) =  (\identity - J_U (\image))^T (\image - U(\image)),
\end{equation}
where the matrix-vector product with $J_U^T(\image)$ can be computed using backpropagation.
Typically, $U$ is realized as an encoder-decoder architecture $U = D \circ E$, for example a UNet \cite{ronneberger2015u} or DRUNet \cite{Drunet2022}.
Such networks are commonly initialized as pretrained image denoising or restoration networks \cite{ZouLiuWoh2023}. 

\Gls{lsr} has been used for denoising \cite{bigdeli2017image}, sparse-view \gls{ct} \cite{LiSch2020}, and, in combination with unrolling, for \gls{mri} reconstruction \cite{aggarwal2018modl}.
More recently, it has been used in the context of convergent \gls{pnp} \cite{cohen2021has,hurault2022gradient} as gradient-step denoiser, and in \cite{ZouLiuWoh2023} using bilevel training.
In \cite{obmann2021augmented}, the regularizer \eqref{eq:LSR} is extended by an additional regularization within the fixed-point set of $U$.

\subsection{Learned Proximal Networks}
\label{sec:LPNs}

In the \gls{pnp} framework, the proximal operator $\prox_\regularizer$ is learned from data.
To this end, we need to parameterize $\prox_\regularizer$, and ensure that the resulting operator is a prox.
\Glsxtrfullpl{lpn} \cite{fang2023s} provide a solution that  allows evaluating the underlying $\regularizer$ even though it is parameterized implicitly.
Based on the observation that gradients of convex functions fully characterize proximal operators \cite{GriNik2020}, we define the reconstructor $\Psi_\theta \colon \R^\dimimage \to \R^\dimimage$ as
\begin{equation}\label{eq:LPN}
    \Psi_\theta(\image) = \prox_\regularizer(\data) = \grad \Phi_\theta(\image),
\end{equation}
where $\Phi_\theta\in C^1(\R^\dimimage)$ is a strongly convex potential with learnable parameters $\theta$.
Importantly, \glspl{lpn} can provide proximals of \textit{nonconvex} regularizers $\regularizer$ since $\regularizer$ is convex if and only if $\Phi_\theta$ is $1$-Lipschitz continuous (see \cite{moreau1965proximite,GriNik2020}), which  is not required in \glspl{lpn}.
The value $\regularizer(\image)$ for some $\image \in \R^\dimimage$ can be recovered via
\begin{equation}\label{eq: implicit reg computation}
	\regularizer(\image) = \bigl\langle\Psi_\parameters^{-1}(\image), \image \bigr\rangle - \frac{1}{2}\| \image \|^2_2 - \Phi_\parameters(\Psi_\parameters^{-1}(\image)),
\end{equation}
and its gradient reads $\grad \regularizer(\image) = \Psi_\theta^{-1}(\image) - \image$.
The inverse $\Psi_\parameters^{-1}(\image)$ satisfies
\begin{equation}\label{eq:ObjectiveExplicitR}
	\Psi_\parameters^{-1}(\image) = \argmin_{\mathbf v \in \R ^d} \{\Phi_\parameters(\mathbf v) - \langle \image, \mathbf v \rangle\},
\end{equation}
allowing for its efficient computation by minimizing this strongly convex objective with the conjugate gradient method.

Typically, \glspl{lpn} are trained on patches of size $l\times l$.
When applying them to images $\image$ of larger size $d=N \times N$, a classical approach is to employ a sliding window method with stride $s$ satisfying $l \bmod s = 0$.
To ensure that this results in a proximal operator, special care is necessary.
We first zero-pad $\image $ on one side of each dimension by $(s - (N \bmod s)) \bmod s$ entries and then add another $l - s$ zeros at both sides.
As a result, all pixels of $\vec x$ are covered by $(l/s)^2$ of the patches when sliding over the padded image $\tilde \image$.
Denoting this (linear) padding process by $\vec P$, the resulting patch-based \gls{lpn} reads
\begin{align}
    \tilde \Psi_\theta(\image) = \frac{s^2}{l^2} \vec P^T \sum_i E_i^T  \Psi_\theta( E_i (\vec P \image)),
\end{align}
which is the gradient of the convex function $\tilde \Phi_\theta(\image) = \frac{s^2}{l^2} \sum_i \Phi_\theta (E_i (\vec P \image))$ and thus again a proximal operator.

In practice, we realize the potential $\Phi_\theta$ by an \gls{icnn} with softplus activation (see Section \ref{sec:ICNN}) and add a small quadratic term to make it strongly convex.
The \gls{icnn} is based on the \gls{cnn} architecture \eqref{eq:feedforward} and is more complex than the configuration described in Section \ref{sec:ICNN}.
We initialize $\parameters$ by taking the exponential of a Gaussian random variable to ensure non-negative values at initialization, which empirically improved the training speed.

Learning proximal or maximally monotone operators for deriving convergent \gls{pnp} algorithms has been previously explored in \cite{HHNPSS2019,HNS2021,hurault2022gradient,PRTW2021} and also the idea of employing gradient-based architectures in the context of \gls{pnp} is not specific to \glspl{lpn}.
In particular, also the so-called \emph{gradient-step} denoisers \cite{cohen2021has,hurault2022gradient} are designed as gradients of a potential $\Phi_\theta$.
The work \cite{HurLec2022} makes use of the results in \cite{Nikolova2015} to note that gradient-step denoisers induce a proximal operator whenever $\grad \Phi_\theta$ is 1-Lipschitz.
Since this is hard to enforce, a relaxation is used in implementations.
Other works have explored weakly-convex regularizers $\regularizer$ \cite{GouNeuUns2023,shumaylov2024weakly}, for which $\prox_\regularizer$ is well-defined.
However, these explicit architectures for $\regularizer$ do not provide closed-form evaluation for $\prox_\regularizer$.

\section{Overview of Training Methods}
\label{sec:training}

\begin{table}[t]
\centering
\caption{Overview of Training Methods.\label{tab:Training}}
\setlength\tabcolsep{1.5pt}
\begin{tabular}{llllll}

\toprule
Data & Approach & Physics & Description \\ 
        
\midrule
supervised & Bilevel Learning (\gls{bl-ift}, \gls{bl-jfb}, \gls{maid}) & \cmarkc/\xmarkc & Section \ref{sec:bilevel} \\%
\midrule
semi-supervised & \glsxtrfull{nett} & \cmarkc/\xmarkc & Section \ref{Sec:NETT}\\
& \glsxtrfull{ar} & \cmarkc/\xmarkc & Section \ref{sec:AR} \\
\midrule
unsupervised & \glsxtrfull{sm} & \xmarkc & Section \ref{sec:ScoreMatching} \\
& \glsxtrfull{patchml} & \xmarkc & Section \ref{sec:ScoreMatching}\\
& \glsxtrfull{pm} & \xmarkc & Section \ref{Sec:ProximalMatching} \\
\bottomrule
\end{tabular}
\end{table}

After choosing a parametric regularizer $\regularizer_\parameters$, the challenge lies in finding suitable parameters $\parameters$.
In Table~\ref{tab:Training}, we summarize the training methods included in our comparison.
They mainly differ in how they incorporate training data and whether they make use of the operator $\forwardop$.
We classify the methods into supervised, semi-supervised or unsupervised as follows.
Supervised methods can include the operator $\forwardop$ and use paired data of ideal images and corrupted data.
Semi-supervised data may also use $\forwardop$ but do not rely on paired data.
Unsupervised methods will not use $\forwardop$ nor any data related to the inverse problem of interest. Some methods can naturally be applied in multiple regimes.

For many approaches, suitable parameters can be found independently of $\forwardop$, allowing deployment to any inverse problem after training.
Regardless of the chosen training method, we deploy $\regularizer_\parameters$ within the variational objective \eqref{eq:VarProb} during evaluation.
This may require a further tuning of e.g., the regularization parameter $\alpha$.
Details on the minimization of \eqref{eq:VarProb} are given in Section~\ref{sec:OptimProb}.

\subsection{Bilevel Learning}\label{sec:bilevel}
Given paired training data $(\image_i,\data_i)\in \R^\dimimage \times \R^\dimdata$, $i=1,\ldots,\ntraining$, supervised bilevel learning considers the nested optimization problem
\begin{align}
    &\min_{\parameters} \biggl\{\upperloss(\parameters) = \frac{1}{\ntraining}\sum_{i=1}^\ntraining
    \upperdist{\hat \image_{\data_i}(\parameters)}{\image_i}\biggr\}\text{ subject to}\label{eq:bilevel-prob}\\
    &\hat \image_{\data_i}(\theta) = \argmin\limits_{\image}\bigl\{\lowerloss_{\data_i}(\image; \parameters) = \datafit(\forwardop\image, \data_i) + \regparam \learnedreg(\image) \bigr\},\label{eq:lower-level}
\end{align}
where $\upperloss$ assesses $\parameters$ by comparing the reconstruction $\hat{\image}_{\data_i}(\parameters)$ to the ground truth image $\image_i$ via the function $G_{\image_i}$. The reconstruction $\hat{\image}_{\data_i}(\parameters)$ is connected to $\learnedreg$ via the lower-level problem \eqref{eq:lower-level}.

\paragraph*{Minimization and Hypergradients}
Most bilevel solvers rely on gradient-based optimization. In the learning context, one often uses stochastic variants such as \gls{adam} \cite{KinJim2015}, which use only a random subset of the training data $(\image_i,\data_i)$ for every update of $\parameters$.
This requires the (stochastic) gradient of $\upperloss$ with respect to $\parameters$, which is often referred to as hypergradient. We now review various ways to compute this gradient. For notational simplicity, we set $\ntraining = 1$ in the following.
Using the chain rule, we obtain
\begin{equation}\label{eq:hypergrad_chain}
\grad_\parameters \upperloss(\parameters) = 
[D_{\parameters}\hat \image_{\data_i}(\parameters)]^T 
\upperdistderiv{\hat \image_{\data_i}(\parameters)}{\image_i},
\end{equation}
where $D_{\parameters}\hat \image_{\data_i}(\parameters)$ is the derivative of $\parameters\mapsto\hat \image_{\data_i}(\parameters)$. The main challenge of \eqref{eq:hypergrad_chain} is the computation of $D_{\parameters}\hat \image_{\data_i}(\parameters)$.
A comparison and detailed discussion of approaches can be found in \cite{BKK2019,JiYanLia2021,LiaXioFet2018,ZucSac2022}. We now briefly discuss four popular options.

\subsubsection{Implicit Differentiation} \label{Sec:IFT}
The \gls{bl-ift} \cite{bengio2000gradient,grazzi2020iteration,samuel2009learning} approach involves implicitly differentiating the optimality condition of \eqref{eq:lower-level}, which reads
\begin{equation}\label{eq:ift1}
    \grad_{\image}
    \lowerloss_{\data_i}(\hat \image_{\data_i}(\parameters);\parameters)=0.
\end{equation}
Given $\parameters$, we compute $\hat \image_{\data_i}(\parameters)$ as described in Section \ref{sec:OptimProb} (also known as forward solve).
If $\lowerloss_{\data_i}(\cdot\ ; \parameters) \in C^2(\R^\dimimage)$ and its Hessian $\vec S(\parameters) = \nabla_{\image}^2\lowerloss_{\data_i}(\hat \image_{\data_i}(\parameters);\parameters)$ is invertible for a given $\parameters$, then the \gls{ift} guarantees the existence of a continuously differentiable solution map $\parameters\mapsto\hat \image_{\data_i}(\parameters)$ locally.
The invertibility condition is for example satisfied if $\lowerloss_{\data_i}(\cdot\ ; \parameters)$ is strictly convex or if $\hat \image_{\data_i}(\parameters)$ is a nondegenerate local minimum.
Differentiating \eqref{eq:ift1} with respect to $\parameters$ leads to
\begin{equation}\label{eq:ift2}
    0=\vec S(\parameters) D_{\parameters}\hat \image_{\data_i}(\parameters) + \vec J(\parameters)
\end{equation}
with $\vec J(\parameters)$ given by the Jacobian of $\parameters\mapsto\nabla_\image\lowerloss_{\data_i}(\hat \image_{\data_i}(\parameters);\parameters)$, that is $\vec J(\parameters) = \nabla_{\parameters,\image}\lowerloss_{\data_i}(\hat \image_{\data_i}(\parameters);\parameters)$.
Thus, the hypergradient \eqref{eq:hypergrad_chain} is  formally given by
\begin{equation}\label{eq:formula_grad}
    \grad_\parameters \upperloss(\parameters)= -[\vec J(\parameters)]^T
    [\vec S(\parameters)]^{-1}
    \upperdistderiv{\hat \image_{\data_i}(\theta)}{\image_i}.
\end{equation}
In practice, we compute the solution $\hat \image_{\data_i}(\theta)$ of \eqref{eq:lower-level} numerically and replace the inversion of the Hessian in \eqref{eq:formula_grad} by solving the linear system \eqref{eq:ift2} for $\hat \image_{\data_i}(\theta)$ with an iterative algorithm (such as \gls{cg} or \gls{minres}).
Assuming that both steps are exact and that the Hessian $\vec S(\parameters)$ is invertible, \eqref{eq:formula_grad} can be computed exactly.
However, iterative solvers only compute an approximation, so the accuracy for both solvers has to be carefully selected.
The computational complexity and memory requirements for solving \eqref{eq:ift2} are independent of the solver specifications for the lower-level problem \eqref{eq:lower-level}.
In particular, we do not have to trace its computational path.
However, solving the linear system \eqref{eq:ift2} requires evaluating many Hessian-vector products, the number of which depends on the system’s conditioning. When computed via automatic differentiation, each iteration can become expensive, especially for $\learnedreg$ with large architectures.

Recent work has extended the \gls{ift} to merely Lipschitz continuous lower-level problems~\cite{BolPau2021}, which is particularly relevant in practice since many approaches employ non-smooth activation functions such as \gls{relu} in $\learnedreg$.

\subsubsection{Unrolling}
Recall that we compute $\hat \image_{\data_i}(\parameters)$ with an iterative forward solver of the form
\begin{equation}\label{eq:IterativeUpdate}
    \image^{(k+1)}=T(\image^{(k)};\parameters), \text{ for }k=0,1,\dots,\nunrolling-1,
\end{equation}
where $\nunrolling$ is the fixed number of iterations, $\image^{(0)}$ is some starting guess and $T(\cdot; \parameters)$ is defined such that solutions of $\eqref{eq:ift1}$ are fixed points of $T(\cdot; \parameters)$.
To simplify our considerations, we only discuss the gradient descent step $T(\image;\parameters) = \image - \tau \grad_{\image}\lowerloss_{\data_i}(\image;\parameters)$, where the step size $\tau$ is added to the learnable parameters $\parameters$.
More sophisticated routines with momentum, such as the evaluation routine from Section \ref{sec:OptimProb}, can be used instead.
In contrast to the \gls{bl-ift} approach, we compute  $D_{\parameters}\hat \image_{\data_i}(\parameters)$ by applying the chain rule to \eqref{eq:IterativeUpdate}, which leads to
\begin{align}\label{eq:GradUnrolling}
     &D_{\parameters}\hat \image_{\data_i}(\parameters) 
     \approx D_\parameters \image^{(\nunrolling)} = \sum_{k=1}^{\nunrolling} \vec A_{\nunrolling}\ldots \vec A_{k+1}\vec B_k ,\text{ where }\\
     &\vec A_{k+1} = D_{\image} T(\image^{(k)};\parameters), \text{ and } \vec B_{k+1} = D_{\parameters} T(\image^{(k)};\parameters).\notag
\end{align}
This approach is commonly known as unrolling \cite{mehmood2020automatic,ochs2016techniques} and corresponds to the \emph{exact} gradient of the finite iterative scheme \eqref{eq:IterativeUpdate}.
An overview of unrolling is provided in \cite{MonLiEld2021}, and a more general perspective based on the \emph{learning-to-optimize} framework is given in \cite{CheCheChe2022}.
The expression \eqref{eq:GradUnrolling} can be efficiently evaluated using backpropagation, which requires storing the intermediate variables $\image^{(k)}$ for $k=1, \dots, \nunrolling$.
Hence, the memory requirement of backpropagation through \eqref{eq:IterativeUpdate} grows linearly with $\nunrolling$, which restricts the number of steps $\nunrolling$ that we can take.
However, if $\nunrolling$ is small, \eqref{eq:IterativeUpdate} does not necessarily lead to a good approximation of a minimizer $\hat \image_{\data_i}(\parameters)$ for \eqref{eq:lower-level}.
Still, unrolling has been used successfully together with an adaption of $\alpha$ in \eqref{eq:VarProb} for the \gls{crr} \cite{GouNeuBoh2022}.
In principle, the memory usage can be decreased by checkpointing \cite{GruMunDan2016}.
To stabilize the training, we can regularize $\nabla_{\image}^2 \lowerloss_{\data_i}(\cdot; \parameters)$ in \eqref{eq:lower-level} \cite{BaiKolKol2021} or perform \eqref{eq:IterativeUpdate} with a random number of steps $\nunrolling$ and random initializations $\image^{(0)}$ \cite{AniPokLia2022}.

\subsubsection{Jacobian-free Backpropagation} \label{Sec:JFB}
A memory-efficient alternative to unrolling is \gls{jfb} \cite{BolPauVai2023,FunHeaLi2022} (also known as truncated backpropagation \cite{ShaCheHat2019,VicMetSoh2022}).
Instead of backpropagating through the whole scheme \eqref{eq:IterativeUpdate}, we only do so for the last $\nbackprop$ steps and approximate
\begin{equation}\label{eq:TruncGrad}
    D_\parameters \image^{(\nunrolling)} \approx \sum_{k=\nunrolling-\nbackprop+1}^{\nunrolling}  \vec A_{\nunrolling}\ldots \vec A_{k+1}\vec B_k.
\end{equation}
Under certain regularity assumptions, the truncation error induced by \eqref{eq:TruncGrad} decays exponentially as $\nbackprop$ increases \cite{ShaCheHat2019}.
Moreover, the obtained hypergradient is a descent direction for \eqref{eq:bilevel-prob} under reasonable assumptions \cite{ShaCheHat2019}.
Since checking the assumptions is infeasible, the results merely serve as motivation.

There is a direct relation to the \gls{bl-ift} approach.
Let $\vec A_{\infty} = D_{\image} T(\hat \image_{\data_i}(\parameters);\parameters)$ with $T$ from \eqref{eq:IterativeUpdate} and $\vec B_{\infty} = D_{\parameters} T(\hat \image_{\data_i}(\parameters);\parameters)$.
Then, close to the equilibrium $\hat \image_{\data_i}(\parameters)$, we can approximate \eqref{eq:TruncGrad} with the Neumann series
\begin{equation}\label{eq:NeumannGrad}
    D_{\parameters}\hat \image_{\data_i}(\parameters)
    \approx  \biggl(\sum_{k=0}^{\nbackprop -1} \vec A_\infty^k\biggr)\vec B_\infty,
\end{equation}
which converges to $(\identity -\vec A_\infty)^{-1}\vec B_\infty$ as $\nbackprop \to \infty$ if $\Vert \vec A_{\infty} \Vert_2 < 1$.
The formula~\eqref{eq:NeumannGrad} is known as Neumann backpropagation~\cite{LiaXioFet2018} and differs from \eqref{eq:TruncGrad} by using only the output $\hat \image_{\data_i}(\theta)$ instead of the last $\nbackprop$ steps of \eqref{eq:IterativeUpdate}.
For gradient descent, we have $\vec A_{\infty} = \identity - \tau \grad^2_{\image} \lowerloss_{\data_i}(\hat \image_{\data_i}(\parameters); \parameters)$.
It holds $\Vert \vec A_{\infty} \Vert_2 < 1$ if $\tau$ is small enough and $\nabla^2_{\image} \lowerloss_{\data_i}(\hat \image_{\data_i}(\parameters);\theta)$ is positive definite, namely if $\hat \image_{\data_i}(\parameters)$ minimizes $\lowerloss_{\data_i}(\cdot;\parameters)$.
In this case, \eqref{eq:NeumannGrad} converges to the \gls{ift}~gradient \eqref{eq:formula_grad} as $\nbackprop \to \infty$.
A deeper study can be found in \cite{LiaXioFet2018}.
In practice, choosing $\nbackprop=1$ often works well \cite{BolPauVai2023,FunHeaLi2022}.

\subsubsection{Adaptive Accuracy} \label{Sec:MAID}
Several works have analyzed the convergence of bilevel optimization based on the assumption that the hypergradients are computed with high accuracy \cite{pedregosa2016hyperparameter}.
Many of these approaches rely on line-search, which requires access to exact solutions of the lower-level problem \eqref{eq:lower-level}. 
However, \eqref{eq:lower-level} is typically solved approximately using an iterative method.
Hence, the inexact solution $\tilde{\image}_{\data}(\parameters_k)$ and $D_{\parameters}\tilde{\image}_{\data}(\parameters_k)$ for the upper level problem \eqref{eq:bilevel-prob} at iteration $k$ might be too inaccurate for the \gls{bl-ift} theory to hold.

To address this, the \glsxtrfull{maid}~\cite{salehi2024adaptively} adaptively selects both the upper-level step size and the accuracy of the lower-level solver.
This requires error bounds for the lower-level solutions and computable error bounds for the inexact hypergradients.
At each upper-level iteration $k$, an approximate lower-level solution  $\tilde{\image}_{\data}(\parameters_k)$ satisfying $\|\tilde{\image}_{\data}(\parameters_k) - \hat{\image}_{\data}(\parameters_k)\|\leq \epsilon_k$ is computed for a given tolerance $\epsilon_k \geq 0$. 
Then, an approximation $\vec q_k$ of $\vec S(\parameters_k)^{-1}\upperdistderiv{\tilde \image_{\data}(\parameters_k)}{\image}$ is computed by solving the corresponding linear system up to the tolerance $\delta_k \geq 0$.
An approximate hypergradient $\vec z_k$ that satisfies $\|\vec z_k - \nabla L(\theta_k)\| = \mathcal{O}(\epsilon_k + \delta_k)$ \cite{ehrhardt2024analyzing} is then given by 
\begin{equation}\label{eq:approx-hypergrad}
    \vec z_k=-\bigl[\nabla_{\parameters,\image}\lowerloss_{\data_i}(\tilde \image_{\data_i}(\parameters);\parameters)\bigr]^T\vec q_k.
\end{equation} 
Motivated by Armijo-type line search, we define the following function for checking sufficient decrease
\begin{equation}
    \label{eq:line-search}
    \xi(\alpha_k)\coloneq 
    \Delta^+_{k+1} - \Delta^-_{k} + \lambda\alpha_k\|\vec z_k\|^2,
\end{equation}
where $\Delta^{\pm}_{l} \coloneq 
\upperdist{\tilde \image_{\data}(\parameters_{l})}{\image}
\pm
\left(
\|\upperdistderiv{\tilde \image_{\data}(\parameters_{l})}{\image}\|
\epsilon_{l}
+
\frac{\gamma}{2}\epsilon_{l}^2\right)$ are upper and lower bounds on $L(\theta_l)$, respectively.
Here, $\gamma$ is the Lipschitz constant of $\nabla G_{\image}$.
If $\xi(\alpha_k) \leq 0$ for some $\lambda>0$ (e.g., $\lambda = 10^{-4}$), sufficient decrease in the objective \eqref{eq:bilevel-prob} can be ensured \cite[Lemma\ 3.5]{salehi2024adaptively}.
Otherwise, the accuracy or the step size is unsuitable and needs to be modified.
The full scheme is given as Algorithm~\ref{alg:MAID}, and its convergence to a critical point is shown in \cite[Thm.\ 3.19]{salehi2024adaptively}.

The implemented \gls{maid} uses gradient descent as a starting point and introduces the inaccurate handling of gradients.
Of course, this idea can be generalized to other algorithms.
However, extensions to stochastic gradients are not straightforward due to its reliance on backtracking.
A stochastic version with nonadaptive accuracy has been recently proposed~\cite{salehi2025ssvm}.

\begin{algorithm*}[t]
\caption{\gls{maid} to solve bilevel learning problem \eqref{eq:bilevel-prob}. \\ Hyperparameters: step size controls $ 0 < \underline{\rho} < 1 < \overline{\rho}$; accuracy controls $0 < \underline{\nu} < 1 <  \overline{\nu}$;  maximum backtracking iterations $b \in \mathbb{N}$.}\label{alg:MAID}
\begin{algorithmic}[1]
\State Input: $\theta_{0}$, accuracies $\epsilon_0, \delta_0>0$, step size $\alpha_{0} >0$.
\For{$k=0, 1, \dots$}
\For{$j = b, b+1, \dots$ }\label{bt_loop}
\State{$\vec z_k \leftarrow$ inexact\_grad($\theta_k, \epsilon_k, \delta_k$)}\Comment{inexact hypergradient using \eqref{eq:approx-hypergrad}} \label{updated_descend_direction}
\For{$i=0,1,\dots,j-1$}\label{inner_loop}
\If{$\xi(\alpha_k) \leq 0$}\Comment{inexact sufficient decrease using \eqref{eq:line-search}}\label{step:descent}
\State{$\theta_{k+1} \leftarrow \theta_k - \alpha_{k} \vec z_k$}\Comment{gradient descent update}\label{gd_update_step}
\State{go to line \ref{increase_params}}\Comment{backtracking successful}
\EndIf
\State{$\alpha_{k} \leftarrow \underline{\rho}\alpha_k$}
\Comment{decrease step size}
\EndFor
\State $\epsilon_k, \delta_k \leftarrow \underline{\nu} \epsilon_k, \underline{\nu} \delta_k \label{BT_decrease} $  \Comment{backtracking failed; needs higher accuracy}
\EndFor
\State{$\epsilon_{k+1}, \delta_{k+1}, \alpha_{k+1} \leftarrow \overline{\nu} \epsilon_{k}, \overline{\nu} \delta_{k}, \overline{\rho} \alpha_{k}$ }\label{increase_params}\Comment{increase parameters}
\EndFor
\end{algorithmic}
\end{algorithm*}

\subsection{Contrastive Learning}\label{Sec:Contrastive}%

Bilevel learning aims to directly learn the reconstruction operator.
Another class of methods tries to learn a regularizer $\regularizer$ by contrasting \enquote{good} and \enquote{bad} images. This approach shares similarities with contrastive learning, which originates from representation learning \cite{becker1992self, chen2020simple}, and has been applied in various fields such as generative modeling \cite{goodfellow2014generative} and latent space embeddings \cite{radford2021learning}.

\subsubsection{Network Tikhonov} \label{Sec:NETT}

The \glsxtrfull{nett} approach \cite{antholzer2019nett,LiSch2020,obmann2021augmented} learns a regularizer of the form $\regularizer = J \circ \Psi_\parameters$, where $J\colon\R^d\rightarrow[0,\infty]$ is a \enquote{distance} functional and $\Psi_\parameters \colon\R^d\to\R^d$ is a parametric network designed to extract artifacts.
A typical choice for $\Psi_\parameters$ is an encoder–decoder architecture $\Psi_\parameters=D_\parameters \circ E_\parameters$, with $J(\cdot)=\Vert\cdot\Vert_2^2$, see also Section~\ref{sec:LSR}.
In particular, $\Psi_\parameters$ aims to model the residual between clean and degraded images, thereby modeling artifacts. 
Given a degradation operator $\mathbf{z}  = G(\image,\data)$, the training of $\Psi_\parameters$ is designed to ensure the following:
(i) when given a degraded image $G(\image,\data)$ with corresponding ground truth $\image$, the network should output the artifacts, i.e., the difference $G(\image,\data)-\image$;
and (ii) when given a clean image $\image$, the network should output zero.
Examples of degradation operators are $G(\image,\data)  = \forwardop^{\dagger}(\forwardop \image + \mathbf{n})$ when the noise model $\mathbf{n}$ and the forward map are known, and $G(\image,\data)  = \forwardop^{\dagger}(\data)$ when supervised training data are available. Both lead to penalization of artifacts introduced by the application of $\forwardop^\dagger$ if \eqref{eq:InvProb} is ill-posed. In any case, the $\regularizer$ is trained such that it takes large values for degraded images $G(\image_i,\data_i)$, $i=1,\dots,n$, and small values for clean images $\image_j, j=1,\dots,m$. This can be achieved with the reconstruction loss
\begin{align}\label{eq:NETTloss}
    \upperloss(\parameters)= 
     \frac{1}{n}\sum_{i=1}^{n}\|G(\image_i,\data_i) - \image_i - \Psi_\parameters(G(\image_i,\data_i))\|^2
       + \frac{1}{m}\sum_{j=1}^{m}\|\Psi_\parameters(\image_j)\|^2.
\end{align}
Note that the training does not involve the functional $J$.
Once trained, the reconstruction is done by solving \eqref{eq:VarProb}, where the regularization parameter $\alpha$ needs to be tuned.
It is also possible to use multiple trained $\regularizer$ to better capture unwanted structure (e.g., nullspace components or noise).
Under natural assumptions on $J$ and $\Psi$, \gls{nett} leads to a convergent regularization method~\cite{LiSch2020}.
A related approach, non-stationary iterated network Tikhonov (iNETT) has been proposed and analyzed in \cite{bianchi2023uniformly}. 
The convergence of the iterated Tikhonov method can be guaranteed by incorporating uniformly convex networks.

\subsubsection{Adversarial Regularization}\label{sec:AR}
\Glsxtrfull{ar} \cite{lunz2018adversarial, MukDit2021,shumaylov2023provably,shumaylov2024weakly, zhang2025learning} operates within a weakly supervised setting.
There, we assume to be given desirable images $\image_i$, $i=1,\ldots,\ntraining$, and (noisy) measurements $\data_j$, $j=1,\ldots,\tilde \ntraining$, which originate from distributions  $\distribution_{X}$ and $\distribution_{Y}$, respectively.
This semi-supervised setting is more realistic in applications where ground truth images are unavailable.
To consider both distributions in the same space, we push-forward $\distribution_{Y}$ from the measurement space $\R^\dimdata$ to the image space $\R^\dimimage$ using a (potentially regularized) pseudo-inverse $\forwardop^\dagger$, giving a distribution $\distribution_{\forwardop} \coloneq (\forwardop^{\dagger})_{\#} \distribution_{Y}$ of images with artifacts.
The key idea of \gls{ar} is to train $\learnedreg$ as a classifier.
More precisely, $\learnedreg$ should be small on real samples from $\distribution_{X}$ and large on the artificial ones from $\distribution_{\forwardop}$.
To achieve this, a natural training loss is
\begin{equation}
\label{eq:AR_loss}
    \upperloss(\parameters) = \frac{1}{\ntraining}
    \sum_{i=1}^\ntraining \learnedreg(\image_i)
    -
    \frac{1}{\tilde \ntraining}
    \sum_{j=1}^{\tilde \ntraining} \learnedreg(\forwardop^{\dagger} \data_j) + \lambda \expected_\image\bigl[\bigl(\left\|\grad \learnedreg(\image)\right\|-1\bigr)_{+}^2\bigr],
\end{equation}
where $\scalarA_+=\max\{\scalarA,0\}$.
The last term serves as regularization that promotes the classifier $\learnedreg$ to be 1-Lipschitz, inspired by the Wasserstein GAN (WGAN) loss \cite{arjovsky2017wasserstein}, and the expectation is taken over points of the form
\begin{equation}\label{eq:ar_gradpen}
   \image = t\image_i+(1-t)\forwardop^\dagger\data_j, 
\end{equation}
where $\image_i\sim\distribution_{X},\ \data_j\sim\distribution_{Y},$ and $t\sim\mathcal{U}[0,1]$. 
A patch-based variant of the \gls{ar} training was proposed in \cite{prost2021learning} under the name \glsxtrfull{lar} in combination with padding-free \glspl{cnn}, see Section~\ref{Sec:PatchBased}. Here, we use a loss similar to \eqref{eq:AR_loss} with patches extracted from both $\image_i$ and $\forwardop^\dagger \data_j$.

Utilizing \eqref{eq:AR_loss} provides an interesting characterization of the optimal $\learnedreg$ \cite{lunz2018adversarial}.
Let us assume that $\distribution_{X}$ is supported on a compact set $\set{M}$ which captures the intuition that images lie in a lower-dimensional non-linear subspace of the original space, see also Section \ref{Sec:NETT}. 
Let $\projection_{\set{M}}$ denote the orthogonal projection onto $\set{M}$.
Further, we assume that $\distribution_{X}$ and $\distribution_{\forwardop}$ satisfy $\left(\projection_{\set{M}}\right)_{\#}\distribution_{\forwardop}=\distribution_{X}$.
This means that the reconstruction artifacts are small enough to allow recovery of the real distribution by simply projecting samples from $\distribution_{\forwardop}$ onto $\set{M}$.
Then, the distance function $\image \mapsto \min_{\vec z \in \set{M}}\|\image-\vec z\|$ is a maximizer of 
\begin{equation}
\label{eq:Wass}
\wasserone(\distribution_{\forwardop},\distribution_{X}) = 
\sup _{f \in \text{1-Lip}} \expected_{\image \sim \distribution_{\forwardop}}[f(\image)]-\expected_{\image \sim \distribution_{X}}[f(\image)],
\end{equation}
which is the formal limit of \eqref{eq:AR_loss} for $\lambda \to \infty$.
For evaluation, we rescale $\regularizer$ to ensure that $\Vert \nabla R(\image)\Vert \leq 1$ on the validation set. 
Regarding the variational problem \eqref{eq:VarProb}, we set $\alpha= \mathbb E_{\vec n \sim \distribution_N} \Vert \forwardop^T \vec n \Vert_2$ provided that the noise distribution $\distribution_N$ is known \cite{lunz2018adversarial}.
This initial estimate of $\alpha$ can be further refined if needed.

\subsection{Distribution Matching}

From the Bayesian viewpoint, the ground truth $\image$ and the observation $\data$ in the inverse problem \eqref{eq:InvProb} are samples from distributions $p_X$ and $p_Y$.
As outlined in Section \ref{sec:introduction}, the solution of the variational problem \eqref{eq:VarProb} corresponds to the \gls{map} estimator of $X$ given $Y=\data$.
In particular, $\regularizer$ is given (up to a constant) by $ \regularizer(\image)\propto-\log(p_X(\image))$ or, equivalently, $p_X$ corresponds to the so-called Gibbs prior $p_X(\image)\propto\exp(-\regularizer(\image))$.
Thus, one can estimate $p_X$ to learn the parameters $\parameters$ of $\regularizer_\parameters$. 
In contrast to all previous methods, the ones discussed here are independent of the operator $\forwardop$, the noise model, and the data term $\datafit$.

\subsubsection{Maximum Likelihood Training}

Let 
$
p_\parameters(\image)=Z_\parameters^{-1}\exp(-\regularizer_\parameters(\image))
$
denote the Gibbs prior with normalizing constant $Z_\parameters= \int_{\mathbb{R}^d}\exp(-\regularizer_\parameters(\image))\mathrm{d}\image$ .
Then, given training samples $\image_1,\ldots,\image_n$ from $p_X$, the parameters $\parameters$ can be learned by computing the maximum likelihood estimator
\begin{equation}\label{eq:ML}
\parameters_\mathrm{ML}=\argmax_\parameters \biggl\{\sum_{i=1}^n \log(p_\parameters(\image_i))\biggr\}=\argmin_\parameters\biggl\{\sum_{i=1}^n \regularizer_\parameters(\image_i)+\log(Z_\parameters)\biggr\}.
\end{equation}
The estimator \eqref{eq:ML} is an empirical estimator of the Kullback--Leibler divergence
\begin{equation}\label{eq:ML-LossCont}
    (p_{X} \mid p_{\parameters})_{\mathrm{KL}} = \mathbb{E}_{\image\sim p_X}\left[\log\Bigl(\frac{p_{X}(\image)}{p_{\parameters}(\image)}\Bigr)\right].
\end{equation}
More precisely, by replacing the empirical sum by an expectation, we obtain that
\begin{equation}
\parameters_\mathrm{ML}\approx\argmin_\parameters \mathbb{E}_{\image\sim p_X}[-\log(p_\parameters(\image))]=\argmin_\parameters \,\,(p_{X} \mid p_{\parameters})_{\mathrm{KL}}. 
\end{equation}
The main difficulty in this approach is the computation of $Z_\theta$.
One possibility is to choose the architecture of $\regularizer_\parameters$ such that $Z_\parameters=1$ is true independently of $\parameters$.
This holds for the Gibbs prior of parametric distributions such as Gaussians \cite{alberti2021learning} (which corresponds to the optimal Tikhonov regularizer), \glspl{gmm} \cite{zoran2011learning}, \glspl{nf} \cite{altekruger2023patchnr,ardizzoneanalyzing,denker2021conditional}, or other generative models.

Unfortunately, such models are often not expressive enough to provide a meaningful approximation of $p_X$. 
Therefore, patch-based architectures like \gls{epll} and \gls{patchnr} approximate the distribution of patches instead, see Section~\ref{Sec:PatchBased} for details.
In this case, we use a \gls{patchml} objective
\begin{align}
\argmin_{\parameters} \expected_{\image \sim p_{Q}} \bigl[ -\log p_{\parameters}(\image) \bigr], \label{eq:patchml}
\end{align}
where the distribution $p_Q$ of $l \times l$ patches is induced by $p_X$~\cite[Lem.\ 3]{altekruger2023patchnr}.

We can rewrite the gradient of the maximum likelihood loss in \eqref{eq:ML-LossCont} as
\begin{align}
\nabla_\parameters\mathbb{E}_{\image\sim p_X}[-\log(p_\parameters(\image_i))]&=\mathbb{E}_{\image\sim p_X}[\nabla_\parameters \regularizer_\parameters(\image)]+\nabla_\parameters \log(Z_\parameters)\notag\\
&=\mathbb{E}_{\image\sim p_X}[\nabla_\parameters\regularizer_\parameters(\image)]-\mathbb{E}_{\image\sim p_\parameters}[\nabla_\parameters\regularizer_\parameters(\image)],\label{eq:gan}
\end{align}
see \cite{hinton2002training} for a detailed derivation of the second step.
In the context of imaging, \eqref{eq:gan} was also studied under the name \enquote{difference-of-expectations objective} in \cite{tan2024unsupervised,ZaKn23MRTEnergy,ZaKo21CTEnergy}.
Still, the computation of the second expectation in \eqref{eq:gan} requires sampling from $p_\parameters$.
This is often realized via Monte Carlo sampling, which makes these methods computationally expensive~\cite{habring2025diffusionabsolutezerolangevin,kuric2025gaussianlatentmachineefficient,tan2024unsupervised}.
The form \eqref{eq:gan} also links the maximum likelihood approach with the contrastive learning methods discussed in Section~\ref{Sec:Contrastive}.
There, $p_\parameters$ is replaced by an \enquote{adversarial distribution} of degraded images. 

\subsubsection{Score Matching}\label{sec:ScoreMatching}

An alternative to estimating $p_X$ is to estimate its gradient $\nabla \log p_X$, which is also known as the Stein score.
However, accessing $\nabla \log p_X$ directly is usually intractable.
Instead, we consider the \emph{noisy} random variable $X_\sigma = X + \sigma \eta$ with $\eta \sim \mathcal N(0,I)$.
Then, the score $s_\sigma=\nabla \log p_{\sigma}$ with $p_{\sigma} \coloneq p_{X_\sigma}$ can be characterized by Tweedie's formula \cite{Mi61,Ef11} as
\begin{equation}
s_\sigma=\argmin_{f}\mathbb{E}_{\image\sim X, \vec n\sim \eta}\bigl[\|f(\image+ \sigma \vec n)-\vec n\|^2\bigr].
\end{equation}
Consequently, we can learn the parameters $\parameters$ such that $\regularizer_\parameters$ approximates $\image\mapsto-\log p_\sigma(\image)$ (up to a constant) by minimizing the \glsxtrfull{sm} loss \cite{Hy05,SoEr19} given by
\begin{equation}\label{eq:score_matching}
\parameters_\mathrm{SM}=\argmin_\parameters \mathbb{E}_{\image \sim p_X, \vec n\sim \eta}\bigl[\|\nabla \regularizer_\parameters(\image+\sigma\vec n)- \vec n\|^2\bigr].
\end{equation}
Similarly to \eqref{eq:ML}, the \gls{sm} loss  in \eqref{eq:score_matching} can be interpreted as a divergence. More precisely, $\parameters_\mathrm{SM}$ minimize the Fisher divergence
\begin{equation}\label{eq:fisher_divergence}
    ( p_{\sigma} \mid p_{\parameters})_{\mathrm{F}} = \int_{\R^\dimimage} p_{\sigma}(\image) \| \nabla \log p_{\sigma}(\image) - \nabla \log  p_{\parameters}(\image) \|^2\,\mathrm{d} \image.
\end{equation}
As an important consequence, we observe that the \gls{sm} loss approximates $p_\sigma$ instead of $p_X$, which introduces a bias that increases for larger $\sigma$.
At the same time, if $\sigma$ is chosen too small, the learned $\regularizer_\parameters$ becomes imprecise in low-density areas of $p_X$ since the influence of these areas vanishes in \eqref{eq:fisher_divergence} as $\sigma\to0$.

While the smoothing bias of $p_\sigma$ often limits the effectiveness of $\regularizer_\parameters$ learned by \gls{sm}, the minimization of \eqref{eq:score_matching} is computationally efficient.
Therefore, \gls{sm} is used as a pretraining step for the bilevel routines described in Section~\ref{sec:bilevel}, see also \cite{ZouLiuWoh2023}.
That is, we first compute the optimal parameters $\parameters_\mathrm{SM}$ for the loss \eqref{eq:score_matching}, and then use $\parameters_\mathrm{SM}$ as an initialization in the bilevel problem \eqref{eq:bilevel-prob}.

\subsubsection{Proximal Matching}
\label{Sec:ProximalMatching}

Finally, we discuss an approach to approximate $-\log p_X$ implicitly via its proximal operator.
The latter is parameterized by some network $\Psi_\parameters$.
Such approaches are widely studied in the context of \gls{pnp} methods \cite{fermanian2022learned,HNS2021,reehorst2018regularization,venkatakrishnan2013plug,Drunet2022}.
Here, we solely consider the case where $\Psi_\parameters=\prox_{\regularizer_\parameters}$ for some underlying $\regularizer_\parameters$.
This is fulfilled, for instance, for the \glspl{lpn} in Section~\ref{sec:LPNs}. Another possibility is the gradient-step denoiser \cite{cohen2021has,hurault2022gradient} provided that the involved potential is $1$-Lipschitz continuous (which is intractable to enforce in practice).

To approximate $\prox_{-\sigma^2\log p_X}$ with $\Psi_\parameters$, we assume that we have training samples $\image$ from $X$ with noisy versions $\image_\sigma$ from $X_\sigma = X + \sigma \eta$ with $\eta \sim \mathcal N(0,I)$.
After choosing a loss function $\ell$, we can learn $\Psi_\parameters$ by solving
\begin{equation}\label{eq:mean_loss}
\hat\parameters=\argmin_\parameters\mathbb{E}_{\image,\image_\sigma}[\ell(\image,\Psi_\parameters(\image_\sigma))].
\end{equation}
Common choices for $\ell$ are the square error $\ell(\image,\vec y)=\|\image-\vec y\|_2^2$, or the absolute error $\ell(\image,\vec y)=\|\image-\vec y\|_1$.

Unfortunately, neither of these choices leads to the desired \gls{map} denoiser in \eqref{eq:mean_loss}: as mentioned in Section \ref{sec:introduction}, choosing the square error leads to the \gls{mmse} estimator, while the absolute error leads to a generalization of medians \cite{hastie2009elements}. 
To enforce that the minimizer of \eqref{eq:mean_loss} leads to $\prox_{-\sigma^2\log p_X}$, the authors in \cite{fang2023s} propose the \gls{pm} loss
\begin{equation} \label{eq:prox_matching}
    \ell_\gamma(\vec{x},\vec{y}) =
1 - \frac{1}{(\pi\gamma^2)^{d/2}}\exp\biggl(-\frac{\|\vec{x}-\vec{y}\|_2^2}{\gamma^2}\biggr), \gamma > 0.
\end{equation}
They show that the denoiser $f^* = \argmin_{f\ \mathrm{measurable}}
\lim_{\gamma \searrow 0} \expected_{\vec{x},\vec{y}} \left[  \ell_\gamma \left( \vec{x},f(\vec{y})\right)\right]$ satisfies $f^* = \prox_{-\sigma^2\log p_X}$ almost everywhere \cite[Thm.\ 3.2]{fang2023s}.
In practice, we minimize \eqref{eq:mean_loss} for $\ell=\ell_\gamma$ with decreasing values of $\gamma$ to approximate the limit $\gamma \to 0$, see also \cite[App.\ G]{fang2023s}.
If the architecture $\Psi_\parameters$ is universal, then one obtains the proximal operator of $\regularizer = -\sigma^2\log p_X$.
Importantly, this denoiser $\Psi_\parameters$---as it reflects the prior $p_X$---can be readily deployed within the variational problem \eqref{eq:VarProb} for other inverse problems using the ADMM algorithm \cite{chan2016plug}.
If $\Psi_\parameters$ is parametrized as \gls{lpn}, then convergence of the resulting \gls{pnp}-ADMM iterations to fixed points can be guaranteed, see \cite{fang2023s} for details.

\section{Set-Up for Comparative Study}
\label{sec:setup-exp}
Now, we describe the setup for our experimental comparison.
We built upon the \texttt{DeepInverse} library \cite{tachella2025deepinv}, and our code is available on \href{https://github.com/johertrich/LearnedRegularizers}{GitHub}.

\subsection{Minimization of the Variational Problem}\label{sec:OptimProb}
For most architectures, we minimize \eqref{eq:VarProb} via the \gls{nmapg}~\cite[Suppl.]{Li2015}.
This method builds upon the ideas of FISTA~\cite{beck2009fast}.
By ensuring sufficient decrease on an auxiliary objective, it guarantees convergence to a stationary point for nonconvex problems while maintaining the optimal convergence rate of $\mathcal O(1/k^2)$ on convex problems.
A backtracking linesearch with Barzilai--Borwein initialization ensures convergence even without knowledge of the Lipschitz constant of the gradient.
The stopping criterion is based on the relative step size.
See \href{https://github.com/johertrich/LearnedRegularizers/blob/main/evaluation/nmAPG.py}{GitHub} for details.
For some architectures, the \gls{nmapg} is infeasible and we use a different optimization method for minimizing \eqref{eq:VarProb}:
\begin{itemize}
\item For the patch-based regularizers $\regularizer$ in \eqref{eq:PatchReg}, the evaluation cost scales linearly with the number of patches $s$, making it prohibitive for high-resolution images.
To address this, we instead evaluate $\regularizer$ only on a random subset of patches per iteration \cite{altekruger2023patchnr}.
For both \gls{epll} and \Gls{patchnr}, we solve the variational problem \eqref{eq:VarProb} using \gls{adam} with a cosine-annealed step size schedule.

\item While \glspl{lpn} provably define a regularizer $R$, the latter is only given implicitly via its proximal mapping.
Therefore, we evaluate \glspl{lpn} with a plug-and-play algorithm based on the alternating direction method of multipliers \cite{chan2016plug}, where the implementation from \texttt{DeepInverse} is used.
\end{itemize}

\subsection{Forward and Noise Models}\label{sec:Forward}
We consider two (inverse) problems: denoising and \gls{ct} reconstruction.
The latter has become one of the most accessible imaging modalities in non-destructive testing, security, and medicine.
For denoising, it holds that $\forwardop = \identity$.
Additionally, the images $\image \in [0,1]^d$ are corrupted by additive Gaussian noise $\vec n$ with standard deviation $\sigma = 0.1$.
Regarding our \gls{ct} reconstruction experiment, recall that a scanner acquires multiple measurements while rotating around an object.
We model a sparse-view setting, where $\forwardop$ is given by the discretized X-ray transform with 60 equispaced angles and a parallel beam geometry.
We use the \texttt{DeepInverse} implementation.
To keep the setup simple, we consider Gaussian noise with $\sigma = 0.7$ instead of more realistic Poisson noise.
For our ground truth images $\image \in [0,1]^d$, the measurement range is between 0 and 400.
In both settings, we use the data-fidelity $\datafit(\forwardop\image,\data)=\frac12\|\forwardop\image-\data\|^2$.
For \gls{ct} reconstruction, taking the pseudo-inverse $\forwardop^\dagger$ is also known as \gls{fbp}.

\subsection{Datasets}\label{sec:Data}
For denoising, we use the BSDS500 dataset \cite{arbelaez_contour_2011,MartinFTM01}, which contains 500 color images of size $481\times 321$ of mixed landscape and portrait orientation.
To simplify the setup, we convert them to grayscale.
The dataset is split into 400 images for training and validation, and the 68 images of the BSD68 set for testing.
Following the literature, the remaining 32 test images are discarded.

For the \gls{ct} reconstruction experiments, we consider the LoDoPaB-CT dataset \cite{LeuSch21lodopabct}.
Its ground truth images of size $362\times362$ are based on reconstructions in the LIDC/IDRI database~\cite{armato2011lung}.
While the original dataset is very large, we use the $3522$ images from the validation set for training, and the $128$ images from the first test batch  for testing.

\subsection{Experiments and Evaluation Metric}\label{subsec:experiments}

Within the setup of Sections \ref{sec:Forward} and \ref{sec:Data}, we conduct three  experiments.

\begin{enumerate}
    \item \textbf{Denoising for Natural Images:} We perform denoising of the BSDS500 dataset.
    In the reconstruction step, we choose the same regularization parameter $\alpha$ in \eqref{eq:VarProb} for training and testing since the setups coincide.
    \item \textbf{Generalization to \gls{ct} Reconstruction:} We evaluate the generalization capability of the trained $\regularizer_\parameters$ from the first experiment.
    To this end, we insert $\tilde\regularizer_{\alpha,s}(\image)=\frac{\alpha}{s^2}\regularizer_\parameters(s\image)
    $ into \eqref{eq:VarProb} and fine-tune the regularization and scaling parameters $\alpha$ and $s$ on the first five images from the training split of the LoDoPaB-CT dataset.
    Instead of a grid search, we use our bilevel training method with hypergradients computed by \gls{ift}.
    \item \textbf{Learned \gls{ct} Reconstruction:} Both domain-specific data (LoDoPaB-CT dataset) and the forward operator $\forwardop$ are available for learning $\regularizer_\parameters$.
    The results are expected to improve upon the ones from the second experiment.
\end{enumerate}
We evaluate all results using the \gls{psnr} defined for the ground truth image $\image$ and reconstruction $\hat \image$ as
\begin{equation}
    \mathrm{\gls{psnr}}(\hat \image,\image)=10\cdot\log_{10}\left(\frac{d^2 r^2}{\|\hat \image-\image\|^2}\right),
\end{equation}
where $d$ is the number of pixels and $r$ is the range of the pixel values.
For the denoising problem, we choose $r=1$.
For the \gls{ct} reconstruction experiments, we choose $r=\max_{i,j}(\image_{ij})$, namely the maximal pixel value in the ground truth image.

\subsection{Architectures}

Below, we specify the configurations of each regularizer in the comparison.
\vspace{.2cm}

\noindent\textbf{\gls{crr} and \gls{wcrr}}  The multiconvolution consists of 3 blocks, where the kernels have size $k = 5$ with $4$, $8$ and $64$ output channels, respectively.
This corresponds to $c=64$ kernels with an effective size of 13. 
For the \gls{crr}, we use the potential $\psi^\beta$ from \eqref{eq:MoreauEnvelope}, and for the \gls{wcrr}, we use the modified potential $\tilde \psi^\beta = \psi^\beta - \psi^1$, where in both cases parameter $\beta$ is learnable.
\vspace{.1cm}

\noindent\textbf{\gls{icnn}} 
We use two convolution layers with no skip connection as shown in \eqref{eq:ICNN}.
The kernels have size $k=5$ with 32 output channels. 
The learnable smoothing parameter $\beta$ for the \gls{relu} is initialized as $100$.
\vspace{.1cm}

\noindent\textbf{\gls{idcnn}} 
The \gls{idcnn} is constructed as the difference of two \glspl{icnn}.
The general architecture of the \glspl{icnn} is the same as described above, where the activation is changed to ELU \cite{clevert2015fast} with $\alpha = 1$ for the \gls{ar} result in Table~\ref{tab:numerics:CT}.
In this particular case, changing the activation significantly improved the reconstruction quality.
\vspace{.1cm}

\noindent\textbf{\gls{epll}}
We employ \glspl{gmm} with ${100, 200, 300, 400}$ components and patches of size $6 \times 6$ and $8 \times 8$.
The exact number of components and patch size are selected based on the validation set.
\vspace{.1cm}

\noindent\textbf{\Gls{patchnr}}
For the \gls{nf} we employ 10 affine coupling layers.
Each layer uses a single three-layer MLP with SiLU activations \cite{hendrycks2016gaussian} and a hidden dimension of $512$.
The output of the MLP is split into two parts to obtain the scale $s$ and translation $t$ in \eqref{eq:INNPass}.
The final layer of the MLP is zero-initialized to ensure that the invertible neural network initially equals the identity.
\vspace{.1cm}

\noindent\textbf{\Gls{cnn}}
We use $6$ convolutional layers with $3 \times 3$ kernels.
The choice of the architecture implicitly defines the patch size as $15 \times 15$, following \cite{prost2021learning}.
As the activation function, we deploy the differentiable SiLU.
\vspace{.1cm}

\noindent\textbf{\gls{tdv}}  We use the \gls{tdv}$_3^3$, which consists of $b=3$ macro-blocks each operating on $a=3$ scales.
We use $c=32$ kernels for the convolution layers.
All kernels of the first layer~$\mathbf{W}$ have zero mean.
\vspace{.1cm}

\noindent\textbf{\gls{lsr}}
We choose $U$ as a DRUNet \cite{Drunet2022} with softmax activation and four scales, where each scale consists of two residual blocks with $32$, $64$, $128$ and $256$ channels.
For the \gls{nett} training, we choose $U$ as a \gls{cnn} consisting of 7 layers with $3  \times 3$ kernels, $64$ hidden channels and an additional residual connection at the end.
The activation function is given by CELU \cite{barron2017continuously} with $\alpha = 10$.
\vspace{.1cm}

\noindent\textbf{\gls{lpn}} 
We realize the network $\Phi_\theta$ from \eqref{eq:LPN} as an \gls{icnn} consisting of $7$ convolution layers with $256$ hidden channels.
Every second convolutional layer has stride $2$ instead of $1$.
After each downsampling, a skip connection injects the (downsampled) input image.
For the activation function, we choose softplus with $\beta=100$.
The \gls{icnn} operates on patches of size $64 \times 64$ and is applied to larger images using a sliding window with stride $32$.

\subsection{Training Methods}
Below, we specify the hyperparameters of the deployed training methods.
We always save the checkpoint with the best validation score.
\vspace{.2cm}

\noindent\textbf{\gls{sm}}
We train using the \gls{sm} loss \eqref{eq:score_matching} with $\sigma=0.03$.
For the deployed Adabelief optimizer \cite{ZhTa20}, the learning rate and the number of epochs depend on $\regularizer_\parameters$.
As post-processing, we fit $\alpha$ and $s$ as detailed in Section~\ref{subsec:experiments}.
The resulting $\parameters_\mathrm{SM}$ is also used as  initialization for the \gls{bl-ift} and \gls{bl-jfb} approaches. 
To initialize the methods in Table~\ref{tab:numerics:CT}, we instead use $\sigma=0.015$ and add weight decay in order to achieve more regularity.
\vspace{.1cm}

\noindent\textbf{\gls{bl-ift}} 
We initialize with the $\parameters_\mathrm{SM}$ generated by the \gls{sm} routine.
Then, we train $\regularizer_\parameters$ with the bilevel loss \eqref{eq:bilevel-prob} and $G_{\image} = \Vert \cdot -\image \Vert_2^2$ using the Adabelief optimizer together with the \gls{ift} to compute the hypergradients.
The accuracies for solving the lower-level problem and the linear system for the hypergradients are both set to $10^{-4}$.
Further, the learning rate and the number of epochs depends on $\regularizer_\parameters$.
To improve stability, we apply Hessian norm regularization every 5 epochs.
\vspace{.1cm}

\noindent\textbf{\gls{maid}}
We adopt an Adagrad update \cite{adagrad,adagrad_nonconvex} with preconditioner $1/\sum_{t=0}^k|z_t|$, where $z_t$ is the approximate hypergradient at a successful iteration $t$ of Algorithm~\ref{alg:MAID}.
The training dataset consists of a fixed number of patches, depending on $\regularizer_\parameters$.
For the hyperparameters in Algorithm~\ref{alg:MAID}, we use \smash{$\underline{\rho} = 0.5$}, $\overline{\rho} = 1.25$, $\underline{\nu} = 0.5$, $\overline{\nu} = 1.05$, a maximum of backtracking iterations $b = 5$, and initial accuracies $\epsilon_0 = \delta_0 = 10^{-1}$.
In contrast to \cite{salehi2024adaptively}, we only check for a decrease in the function value in Step \ref{step:descent} of Algorithm~\ref{alg:MAID}.
\vspace{.1cm}

\noindent\textbf{\gls{bl-jfb}}
We adopt the \gls{jfb} approach \eqref{eq:TruncGrad} with $\nbackprop=1$ to compute the hypergradients.
Everything else remains the same as for \gls{bl-ift}. 
\vspace{.1cm}

\noindent\textbf{\gls{nett}}
We use the \gls{adam} optimizer with a learning rate of $10^{-4}$ to minimize \eqref{eq:NETTloss}.
After the training, we fit $\alpha$ as detailed in Section~\ref{subsec:experiments}.
\vspace{.1cm}

\noindent\textbf{\gls{ar}/\gls{lar}}
We minimize the \gls{ar} objective \eqref{eq:AR_loss} using \gls{adam}.
To reduce the computational burden, we actually train on patches, which amounts to \gls{lar}.
The patch and batch size, learning rate, decay rate, and epochs depend on $\regularizer_\parameters$.
After the training, we finetune $\alpha$ and $s$ as detailed in Section~\ref{subsec:experiments}.
\vspace{.1cm}

\noindent\textbf{\gls{patchml}} To minimize the \gls{patchml} objective \eqref{eq:patchml}, we use the \gls{em} algorithm as implemented in \texttt{DeepInverse} for \gls{epll} and gradient-based optimization with \gls{adam} for \gls{patchnr}.
After training, we fit the regularization parameter $\alpha$ using a grid search.
\vspace{.1cm}

\noindent\textbf{\Gls{pm}}
We pretrain the model with the $\ell_1$ loss for 160k iterations with a learning rate of $\num{1e - 3}$.
Then, we continue training with the \gls{pm} loss \eqref{eq:prox_matching} for another 160k iterations using a learning rate of $\num{1e - 4}$.
The parameter $\gamma$ is initialized as $1.28\sqrt{n}$ and reduced by half every 40k iterations, where $n$ is the data dimension.
To ensure that the \gls{icnn} (which parameterizes the \gls{lpn}) remains convex, we clip its weights after each training step to ensure their non-negativity.
The batch size is 64 for BSDS500 and 128 for the LoDoPaB-CT dataset.

\subsection{Baseline Methods}

In addition to the learned regularizers, we report the \gls{psnr} values for some common baselines.
We use the regularization parameter or the checkpoint with the lowest \gls{mse} on the validation set.

\noindent\textbf{\gls{tv}}
We consider the variational problem \eqref{eq:VarProb} with $\regularizer$ chosen as the anisotropic \gls{tv} \cite{rudin1992nonlinear}.
To minimize \eqref{eq:VarProb}, we employ the \gls{pdhgm} algorithm \cite{chambolle2011first,pock2009algorithm}.
For denoising, the reconstruction \gls{psnr} is $27.30$~dB, and for the \gls{ct} setup, we achieve a \gls{psnr} of $30.99$~dB.
\vspace{.1cm}

\noindent \textbf{DRUNet} \cite{Drunet2022} We use the implementation and weights from \texttt{DeepInverse}, which achieves a reconstruction \gls{psnr} of $29.41$~dB.
Both the training set and the model size ($32.6$M parameters) are significantly larger than any of the regularizers that we test in this work, see also Table~\ref{tab:bsd68_denoising}.
\vspace{.1cm}

\noindent{\textbf{\gls{fbp}+UNet}}
Our UNet-based postprocessing \cite{jin2017deep}  for \gls{ct} is implemented using \texttt{DeepInverse}.
In this approach, the output of the \gls{fbp} is inserted into a UNet.
The latter is trained to remove artifacts by minimizing the \gls{mse} against the ground truth images.
The UNet uses $5$ scales and has approximately $34.5$M parameters.
We train with \gls{adam} for 100 epochs with a learning rate of $\num{1e - 3}$.
This leads to a reconstruction \gls{psnr} of $33.03$~dB. 
\vspace{.1cm}

\noindent\textbf{\Gls{lpd}}
This \gls{ct} reconstruction method \cite{adler2018learned} unrolls the \gls{pdhgm} for a fixed number of steps and replaces the proximal operators with \glspl{cnn}.
We use $6$ steps and implement the networks for the dual variables as small \glspl{cnn}, whereas the networks in the primal space are UNets.
This results in roughly $1$M parameters.
As in \cite{hauptmann2020multi}, we replace the adjoint $\forwardop^T$ in \gls{lpd} with the \gls{fbp}.
The model is trained for $100$ epochs using an initial learning rate of $\num{1 e -4}$ with a cosine decay to $\num{1 e - 6}$.
This leads to a reconstruction \gls{psnr} of $33.71$~dB. 
\vspace{.1cm}

\section{Numerical Results}
\label{sec:results}
\subsection{Experiment 1: Denoising}

The regularizers are trained and evaluated for denoising of natural images. 
Quantitative results are reported in Table~\ref{tab:bsd68_denoising}. 
Some training methods work only with specific regularizers, so not all combinations are compatible.
These fields are grayed out.
Additionally, we use a hyphen for fields that could be filled out, but where the run was computationally intractable or unstable.

The convex models (\gls{crr} and \gls{icnn}) behave similarly, and their performance is largely unaffected by the chosen training scheme.
In particular, the unsupervised  \gls{sm} and the semi-supervised \gls{ar} yield similar results as the supervised bilevel training.
To some extent, this is also true for \gls{wcrr}.
The nonconvex \gls{wcrr}, \gls{idcnn}, \gls{cnn} and \gls{lpn} lead to similar \gls{psnr} values.
The patch-based regularizers (\Gls{epll} and \gls{patchnr}) are not competitive.
The best results are achieved by \gls{tdv} and \gls{lsr} if trained via bilevel learning.
Here, the specific routine (\gls{bl-ift}, \gls{bl-jfb}, \gls{maid}) has negligible impact.
With the other training routines, the performance of the nonconvex architectures   degrades heavily.

Qualitative results are shown in Figure~\ref{fig:denoise_bsd68}.
The trend is similar to Table~\ref{tab:bsd68_denoising}, in that convex models lead to the worst image quality and architectures with the highest number of parameters (\gls{tdv} and \gls{lsr}) lead to the best image quality overall.
Interestingly, the nonconvex models reconstruct the geometry of the large window wrongly.
In particular, \gls{tdv} and \gls{lsr} produce visually appealing windows that differ significantly from the ground truth (see Figure~\ref{fig:denoise_bsd68}).

Figure~\ref{fig:denoise_bsd68:TDV} illustrates the influence of the training scheme, exemplified here for the \gls{tdv} regularizer.
The bilevel training yields good approximations of the ground truth, comparable to those of DRUNet.
\Gls{ar} leads to a smoother image with fewer details and \gls{sm} yields a cartoonish looking image with sharp edges.

\subsection{Experiment 2: Generalization to CT}

We investigate how the models from Experiment 1 generalize to \gls{ct} reconstruction of medical images. 
Quantitative results can be found in Table~\ref{tab:numerics:domainshift}.
As there was no significant difference between the bilevel methods in Experiment 1, we report results only for \gls{bl-jfb}. 
Overall, there are similar trends as for Table~\ref{tab:bsd68_denoising}.
Almost all methods perform better than \gls{tv} even though they have not been trained on the underlying data. 
This means that they generalize fairly well.

Reconstructions can be found in Figure~\ref{fig:ct_lodopab:transfer}.
Here, none of the variational methods yields sharp images comparable to the ground truth.
This is in contrast to the supervised baselines (\gls{fbp}+UNet and \gls{lpd}), which give sharper images with higher \gls{psnr}.
The patch-based regularizers (\gls{epll} and \gls{patchnr})  yield performance comparable to that of \gls{crr} and \gls{icnn}.
This contrasts the results in Table~\ref{tab:bsd68_denoising} even though the same weights are used for both experiments.

\subsection{Experiment 3: CT-specific Training}

The regularizers are trained and evaluated on \gls{ct} reconstruction of medical images.
Due to computational reasons, we only consider the \gls{bl-jfb} mode of bilevel learning. 
Quantitative results are reported in Table~\ref{tab:numerics:CT}.
Overall, similar observations can be made as before.
For \gls{bl-jfb}, the convex  models (\gls{crr} and \gls{icnn}) already perform significantly better than \gls{tv}.
The performance improves slightly for the fairly simple nonconvex and patch-based models \gls{wcrr}, \gls{idcnn}, \gls{epll} and \gls{patchnr}.
The high-parametric models \gls{tdv} and \gls{lsr} achieve the highest \gls{psnr}.
For \gls{ar} training, the difference is much smaller, with convex models and \gls{idcnn} performing slightly worse than other architectures.
As before, the training scheme is important for the more complex architectures.

The quantitative results are visually confirmed in Figure~\ref{fig:ct_lodopab}.
\gls{tdv} and \gls{lsr} give well-defined structures with crisp details that are fairly consistent with the ground truth.
\gls{lsr} achieves the best \gls{psnr} for this image, in line with the results on the whole dataset.
Note that \gls{tdv} and \gls{lsr} are on par with the \gls{lpd} baseline.
In summary, the task-specific training greatly enhances the visual reconstruction quality compared to Figure \ref{fig:ct_lodopab:transfer}.
Figure~\ref{fig:ct_lodopab:comparetraining} shows visual results for different training methods applied to \gls{lsr}, highlighting the outcomes of Experiments 2 and 3 using  \gls{bl-jfb} and \gls{nett}.

\subsection{Training Times for Experiment 1}
So far, we only highlighted the image quality, both quantitatively and qualitatively, after the training has been completed.
Now, we analyze the computational load incurred during training.
To ensure comparability, all experiments are conducted on a single NVIDIA GeForce RTX 4090 GPU with 24 GB memory.
In Table~\ref{tab:bsd68_denoising_training_time}, we report the training times for Experiment 1 in hours for all methods.
In all cases, the models were trained until the loss stabilized.
There are some trends to be observed.
First, simpler models like \gls{crr}, \gls{wcrr}, and to a certain extent \gls{icnn}, required the shortest training time, mostly under an hour.
In contrast, more complex architectures like \gls{idcnn}, \gls{cnn}, \gls{tdv} and \gls{lsr} required significantly longer training time, ranging from several hours to up to 2 days.
Furthermore, \gls{sm} tends to be cheaper than \gls{ar}, which in turn tends to be cheaper than bilevel learning.
In all experiments, \gls{bl-jfb} was consistently faster than \gls{bl-ift} while leading to similar results.

\begin{figure}
\includegraphics[width=\linewidth]{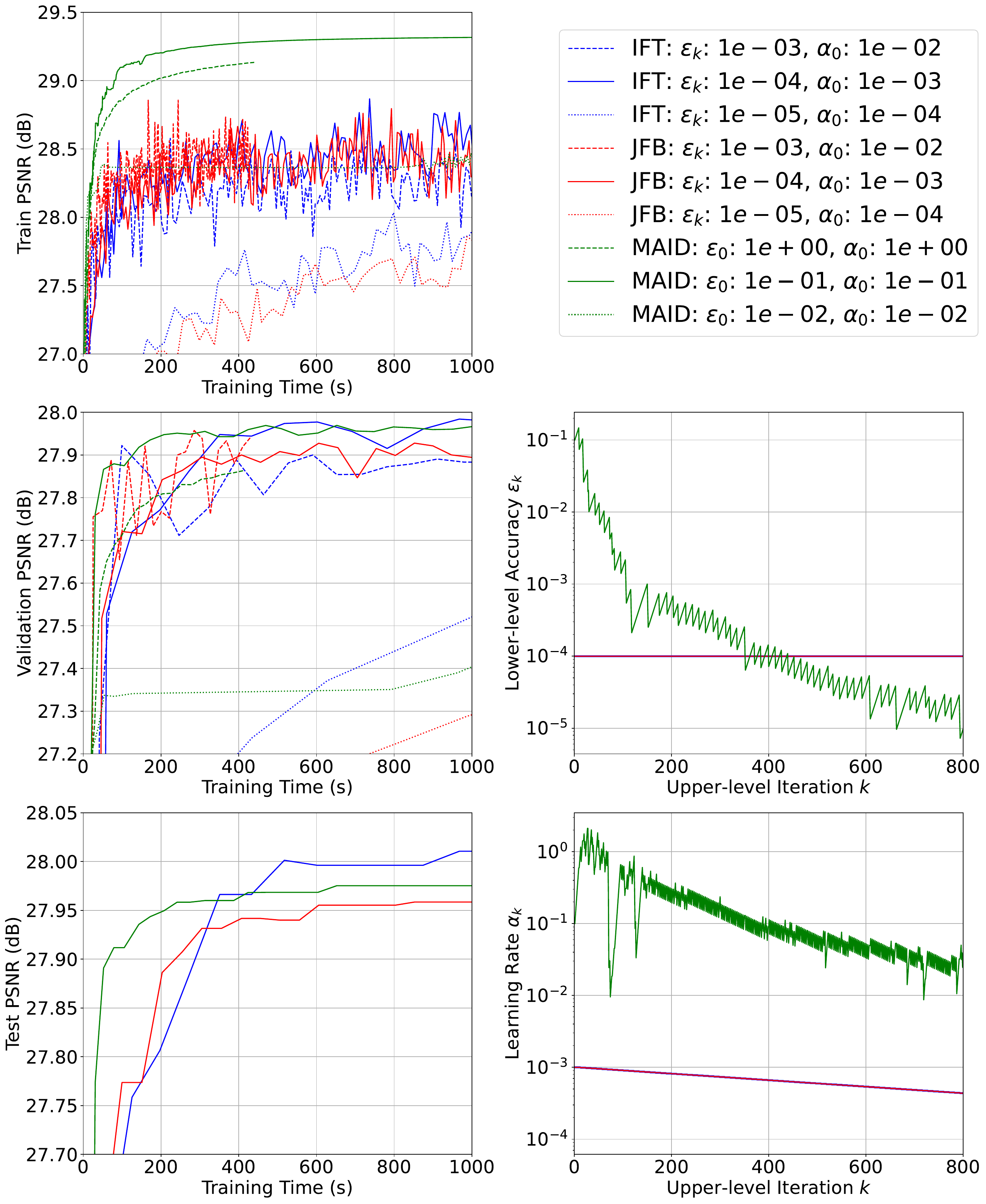}
      \caption{Training time comparison between \gls{maid}, \gls{bl-ift} and \gls{bl-jfb} for \gls{crr}. All algorithms converge to a very similar test \gls{psnr} but vary in speed.}
   \label{fig:maid_comp}
\end{figure}

For bilevel training, we implemented three algorithms: \gls{bl-ift}, \gls{bl-jfb} and \gls{maid}.
These differ in how they compute hypergradients and how they treat solver tolerances and step sizes.
Figure~\ref{fig:maid_comp} compares them for training a \gls{crr}.
For this, we drop the \gls{sm} pretraining since this already leads to nearly optimal parameters (Table \ref{tab:bsd68_denoising}).
In the top row, we see the training and validation \gls{psnr}.
From the training graph, it is clearly visible that \gls{bl-ift} and \gls{bl-jfb} are stochastic methods, whereas the implemented \gls{maid} version is nonstochastic.
We tuned the hyperparameters to maximize the validation \gls{psnr}.
The respective test performances are compared in the bottom row.
All three approaches converge to a similar \gls{psnr}, with \gls{bl-ift} being best.
Interestingly, \gls{maid} reaches a near-optimal \gls{psnr} already after around 200 seconds, which is almost twice as fast as the other methods.
Potential explanations are the lower level accuracy in early iterations, and the much larger step sizes (on average 100 times larger).

\subsection{Evaluation Times for Experiment 1}

Finally, we compare the reconstruction times when solving the variational problem \eqref{eq:VarProb}. 
Again, we use a single NVIDIA GeForce RTX 4090 GPU with 24 GB memory.
Table~\ref{tab:bsd68_denoising_time} reports the average runtime of \gls{nmapg} in seconds until the convergence criterion is reached.
For most combinations of architecture and training scheme, the runtime is below 1 second.
The \gls{crr}, \gls{wcrr}, \gls{icnn}, \gls{idcnn} are faster than the others with maximal speed of around 0.1 seconds reached by \gls{icnn} when trained via \gls{sm} or \gls{ar}. 
More complex architectures tend to need longer with runtimes of 1-2 seconds. 
The ratio between the slowest and the fastest method is around 50.
Due to the many overlapping patches, both \gls{epll} and \gls{patchnr} have a slow algorithmic performance.
Note that their evaluation could be accelerated by stochastic optimization methods or by using the half-quadratic splitting scheme, which was originally used for \gls{epll} \cite{parameswaran2018accelerating, zoran2011learning}.

\newcommand{\WDW}{\cellcolor{gray!10}}
\newcommand{\ILO}{-}
\newcommand{\final}[1]{#1}
\newcommand{\prelim}[1]{#1}

\begin{table}%
\centering
\caption{Experiment 1: \gls{psnr} (in dB) for denoising results on BSD68 with $\sigma=0.1$.
All models are trained for denoising on BSDS500.
\gls{tv}-denoising leads to a \gls{psnr} of 27.3 and a DRUNet reconstruction to 29.41.
The classic \gls{epll} with half-quadratic splitting gave a \gls{psnr} of 28.46. \label{tab:bsd68_denoising}}
\setlength\tabcolsep{1.5pt}
{
\begin{tabular}{cl ccccc ccccc}
\toprule
& Architecture:  
& \gls{crr} 
& \gls{icnn} 
& \gls{wcrr}  
& \gls{idcnn} 
& \gls{cnn} 
& \gls{tdv} 
& \gls{lsr} 
& \gls{epll} 
& \gls{patchnr} 
& \gls{lpn}\\ 
        
\midrule
\multirow{10}{1em}{\rotatebox{90}{Training Scheme}} 
& \gls{bl-ift} 
& 28.01 
& 27.90 
& 28.60 
& 28.58 
& \ILO 
& 29.24 
& 29.25 
& \WDW 
& \WDW 
& \WDW 
\\ 
& \gls{bl-jfb} 
& 28.00 
& 27.89 
& 28.59 
& 28.57 
& 28.89 
& 29.24 
& 29.27 
& \WDW 
& \WDW 
& \WDW 
\\ 
& \gls{maid}
& 28.01 
& 27.82 
& 28.54 
& \ILO
& \ILO
& \ILO 
& \ILO
& \WDW 
& \WDW 
& \WDW 
\\ 
\cmidrule{2-12}
& \gls{ar}/\gls{lar}
& 27.96 
& 27.77 
& 28.48 
& 28.20 
& 28.34 
& 28.62 
& \ILO 
& \WDW 
& \WDW 
& \WDW 
\\ 
& \gls{nett}
& \WDW 
& \WDW 
& \WDW 
& \WDW 
& \WDW 
& \WDW 
& 27.09 
& \WDW 
& \WDW 
& \WDW 
\\
\cmidrule{2-12}
& \gls{sm} 
& 27.94 
& 27.73 
& 28.48 
& 27.93 
& 27.59 
& 27.96 
& 27.61
& \WDW 
& \WDW 
& \WDW 
\\
& \gls{patchml}
& \WDW 
& \WDW 
& \WDW 
& \WDW
& \WDW 
& \WDW
& \WDW
& \final{27.46}
& \final{27.74} 
& \WDW
\\
& \gls{pm} 
& \WDW 
& \WDW 
& \WDW 
& \WDW
& \WDW
& \WDW
& \WDW 
& \WDW
& \WDW 
& 28.33 
\\

\bottomrule
\\
\end{tabular}
}
\vspace{1cm}

\centering
\caption{Experiment 2: \gls{psnr} (in dB) for \gls{ct} reconstruction on the LoDoPaB-CT data set.
All models are trained on BSDS500 \textbf{without} using the operator $\forwardop$.
The \gls{tv} reconstruction achieved a \gls{psnr}  of 30.99 and \gls{fbp} of 19.98.
For BL we report the best-performing instance from Experiment 1.
\label{tab:numerics:domainshift}}
\setlength\tabcolsep{1.5pt}
{
\begin{tabular}{cl ccccc ccccc}
\toprule
& Architecture:
& \gls{crr} 
& \gls{icnn} 
& \gls{wcrr} 
& \gls{idcnn} 
& \gls{cnn} 
& \gls{tdv} 
& \gls{lsr} 
& \gls{epll} 
& \gls{patchnr} 
& \gls{lpn} 
\\ 
        
\midrule
\multirow{8}{1em}{\rotatebox{90}{Training Scheme}} 
& BL (best)
& 32.17 
& 31.99 
& 32.65 
& 32.45 
& 32.69 
& 33.23 
& 33.11 
& \WDW 
& \WDW 
& \WDW 
\\ 
\cmidrule{2-12}
& \gls{ar}/\gls{lar}
& 32.14 
& 31.94 
& 32.61 
& 31.98 
& 32.04 
& 32.43
& \ILO
& \WDW 
& \WDW 
& \WDW 
\\ 
& \gls{nett}
& \WDW 
& \WDW 
& \WDW 
& \WDW 
& \WDW 
& \WDW 
& 30.64 
& \WDW 
& \WDW
& \WDW 
\\
\cmidrule{2-12}
& \gls{sm}
& 32.12 
& 31.85 
& 32.32 
& 31.76 
& 30.03 
& 32.32
& 30.26
& \WDW 
& \WDW 
& \WDW
\\
& \gls{patchml}
& \WDW 
& \WDW 
& \WDW 
& \WDW 
& \WDW 
& \WDW
& \WDW
& 31.94 
& 32.17 
& \WDW 
\\
& \gls{pm} 
& \WDW
& \WDW
& \WDW
& \WDW
& \WDW
& \WDW
& \WDW 
& \WDW
& \WDW 
& 31.29 
\\
\bottomrule
\\
\end{tabular}}
\vspace{1cm}

\centering
\caption{Experiment 3: \gls{psnr} (in dB) for \gls{ct} reconstruction on LoDoPaB-CT data set.
All models are trained on LoDoPaB-CT images using the operator $\forwardop$.
The \gls{tv} reconstruction achieved a \gls{psnr} of 30.99 and \gls{fbp} of 19.98.
The learned \gls{fbp}+UNet achieved a \gls{psnr} of 33.03 and \gls{lpd} of 33.71.
\label{tab:numerics:CT}}
\setlength\tabcolsep{1.5pt}
{
\begin{tabular}{cl ccccc ccccc}
\toprule
& Architecture: & \gls{crr} & \gls{icnn} & \gls{wcrr} & \gls{idcnn} &  \gls{cnn} & \gls{tdv} & \gls{lsr} & 
\gls{epll} & \gls{patchnr} &
\gls{lpn} \\ 
\midrule 
\multirow{7}{1em}{\rotatebox{90}{Training Scheme}} 
& \gls{bl-jfb} 
& \prelim{32.30}  
& \prelim{32.16}  
& \prelim{32.85}  
& \prelim{32.56}
& \ILO 
& \prelim{33.67}
& \prelim{33.72} 
& \WDW 
& \WDW
& \WDW
\\
\cmidrule{2-12}
& \gls{ar}/\gls{lar}
& \prelim{32.23}  
& \prelim{31.98}  
& \prelim{32.48}  
& \final{31.93}
& \final{32.29}
& \prelim{32.33}
& \ILO 
& \WDW 
& \WDW
& \WDW
 \\
& \gls{nett}
& \WDW  
& \WDW  
& \WDW  
& \WDW 
& \WDW
& \WDW 
& \prelim{32.01}
& \WDW 
& \WDW 
& \WDW\\ 
 \cmidrule{2-12}
& \gls{patchml}
& \WDW  
& \WDW  
& \WDW  
& \WDW
& \WDW
& \WDW
& \WDW
& \final{32.55}
& \prelim{32.63}
& \WDW
\\
& \gls{pm} 
& \WDW   
& \WDW   
& \WDW   
& \WDW  
& \WDW 
& \WDW  
& \WDW  
& \WDW  
& \WDW 
& 32.08 
\\
\bottomrule
\end{tabular}}
\end{table}

{
\newlength{\picheight}
\setlength{\picheight}{4.4cm}
\newlength{\cropheight}
\setlength{\cropheight}{2.6cm}
\hypersetup{linkcolor=white}
\newcommand{\addfigure}[3]{%
    \begin{tikzpicture}%
    \node[anchor=south west,inner sep=0] (image) at (0,0) {\includegraphics[height=\picheight]{img/denoise/castle/#2.png}};%
    \begin{scope}[x={(image.south east)},y={(image.north west)}]%
    \node[text=white, anchor=north east] at (0.99,0.99) {\footnotesize \textbf{#3}};
    \node[text=white, anchor=south west] at (0.01,0.01) {\footnotesize \textbf{#1}};%
    \end{scope}%
    \end{tikzpicture}%
}
\newcommand{\addcrop}[3]{%
    \begin{tikzpicture}%
    \node[anchor=south west,inner sep=0] (image) at (0,0) {\includegraphics[height=\cropheight,trim={200 310 60 110}, clip]{img/denoise/castle/#2.png}};%
    \begin{scope}[x={(image.south east)},y={(image.north west)}]%
    \node[text=white, anchor=north east] at (0.99,0.99) {\footnotesize \textbf{#3}};
    \node[text=white, anchor=south west] at (0.01,0.01) {\footnotesize \textbf{#1\phantom{fg}}};%
    \end{scope}%
    \end{tikzpicture}%
}
\newcommand{\addlabel}[2]{%
\rotatebox{90}{\parbox[c]{#2}{\centering #1}\hspace*{-#2}}}
\begin{figure}
    \centering
    \begin{minipage}{\textwidth}
    \addlabel{Data}{\picheight}
    \addfigure{Ground Truth}{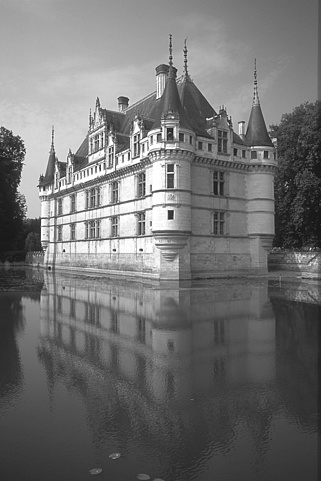}{$\infty$}
    \addfigure{Noisy Data}{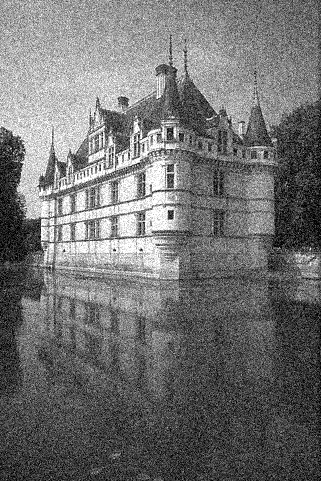}{20.01}\\[-3mm]

    \addlabel{Baseline}{\cropheight}
    \addcrop{Ground Truth}{castle_clean}{$\infty$}
    \addcrop{Noisy Data}{castle_noisy}{20.01}
    \addcrop{\gls{tv}}{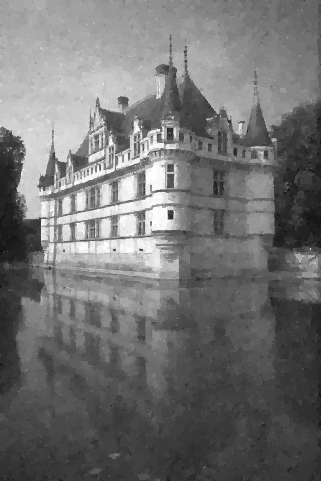}{27.89}
    \addcrop{DRUNet}{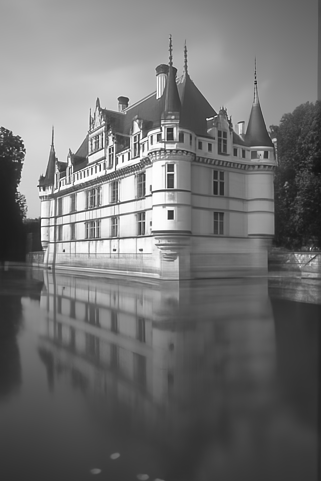}{30.48}\\[-3mm]

    \addlabel{Convex}{\cropheight}
    \addcrop{\gls{crr}}{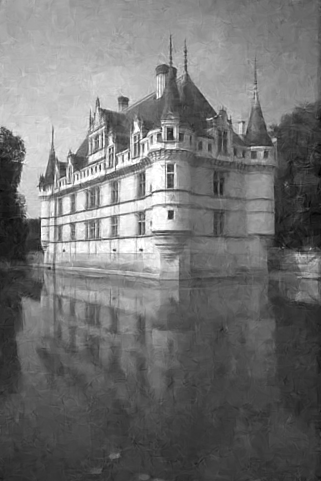}{28.60}
    \addcrop{\gls{icnn}}{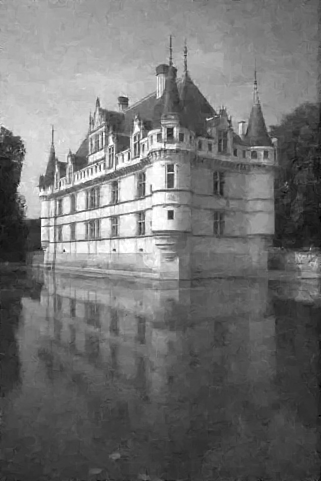}{28.50}\\[-3mm]

    \addlabel{\phantom{Nonconvex}}{\cropheight}    
    \addcrop{\gls{wcrr}}{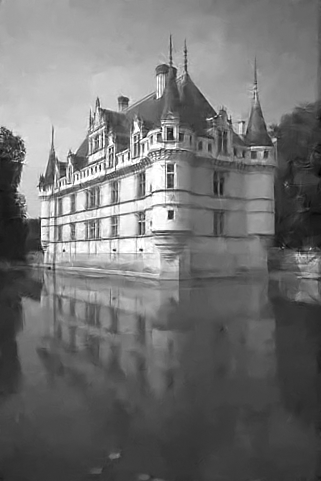}{29.39}
    \addcrop{\gls{idcnn}}{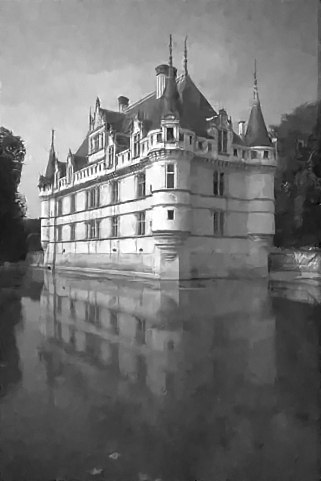}{29.40}
    \addcrop{\gls{epll}}{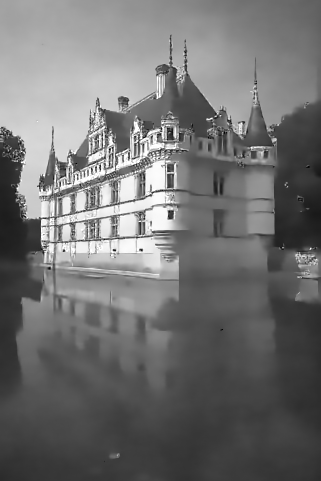}{28.51}
    \addcrop{\gls{patchnr}}{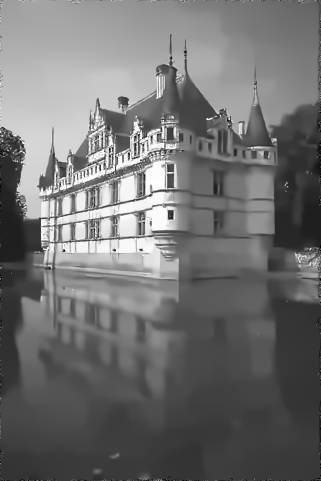}{28.82} \\[-3mm]
    
    \addlabel{Nonconvex}{5cm}
    \addcrop{\gls{cnn}}{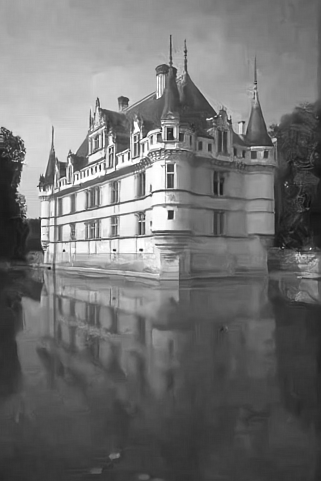}{29.70}
    \addcrop{\gls{tdv}}{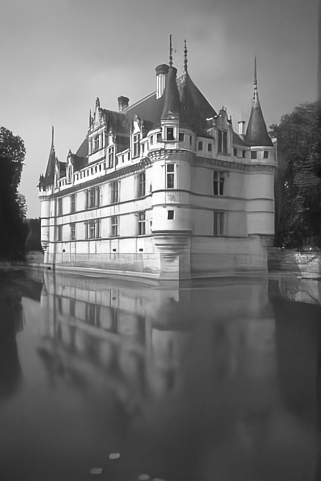}{30.14}
    \addcrop{\gls{lsr}}{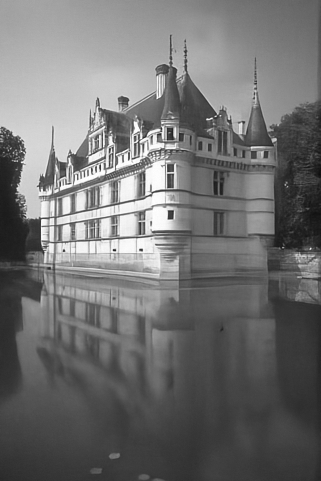}{30.18}
    \addcrop{\gls{lpn}}{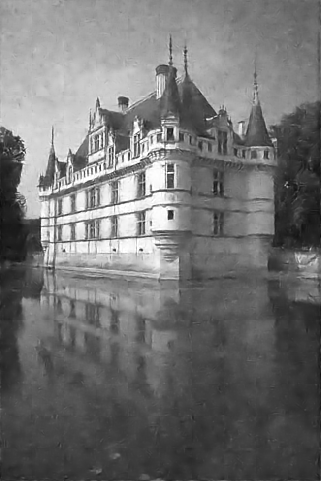}{29.07}
    \end{minipage}
    \caption{Denoising result for Experiment 1 on test image 'castle' from BSD68.}
   \label{fig:denoise_bsd68}
\end{figure}
}

{
\setlength{\picheight}{2.37cm}
\newcommand{\addfigure}[3]{%
    {\hypersetup{linkcolor=black}%
    \begin{tikzpicture}%
    \node[anchor=south west,inner sep=0] (image) at (0,0) {\includegraphics[height=\picheight]{img/denoise/lion/#2.png}};%
    \begin{scope}[x={(image.south east)},y={(image.north west)}]%
    \node[text=black, anchor=north east] at (0.99,0.99) {\footnotesize \textbf{#3}};
    \node[text=black, anchor=south west] at (0.01,0.01) {\footnotesize \textbf{#1}};%
    \end{scope}%
    \end{tikzpicture}%
    }%
}
\newcommand{\addcrop}[3]{%
    {\hypersetup{linkcolor=white}%
    \begin{tikzpicture}%
    \node[anchor=south west,inner sep=0] (image) at (0,0) {\includegraphics[height=\picheight,trim={140 50 220 160}, clip]{img/denoise/lion/#2.png}};%
    \begin{scope}[x={(image.south east)},y={(image.north west)}]%
    \node[text=black, anchor=north east] at (0.99,0.99) {\footnotesize \textbf{#3}};
    \node[text=black, anchor=south west] at (0.01,0.01) {\footnotesize \textbf{#1\phantom{fg}}};%
    \end{scope}%
    \end{tikzpicture}%
    }%
}
\newcommand{\addlabel}[1]{%
\rotatebox{90}{\parbox[c]{\picheight}{\centering
    #1}}
}
\begin{figure}
    \begingroup
    \renewcommand*{\glstextformat}[1]{\textcolor{black}{#1}}%
    \centering
    \begin{minipage}{\textwidth}
    \addlabel{Data}
    \addfigure{Ground Truth}{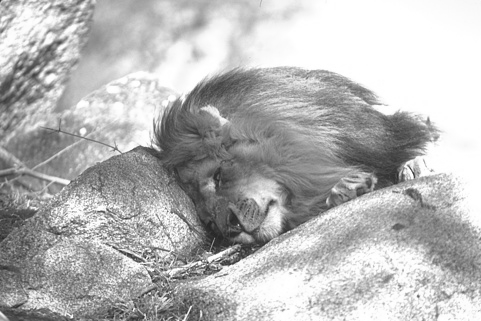}{$\infty$}
    \addfigure{Noisy}{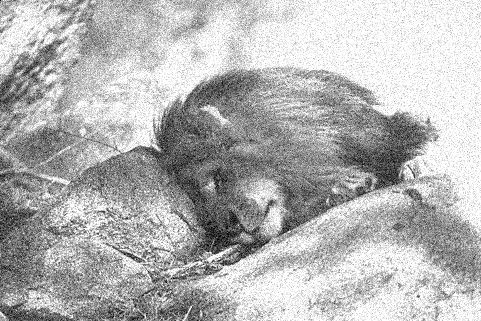}{20.01}
    \\[-3mm]

    \addlabel{Baseline}
    \addcrop{Ground Truth}{lion_gt}{$\infty$}
    \addcrop{Noisy}{lion_noisy}{20.01}
    \addcrop{\gls{tv}}{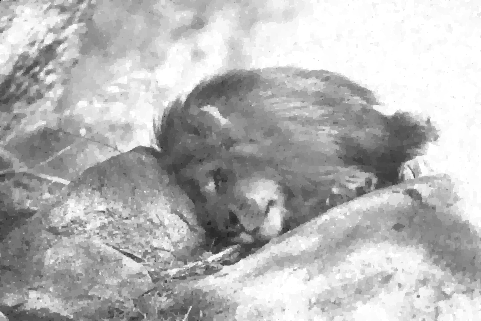}{26.12}
    \addcrop{DRUNet}{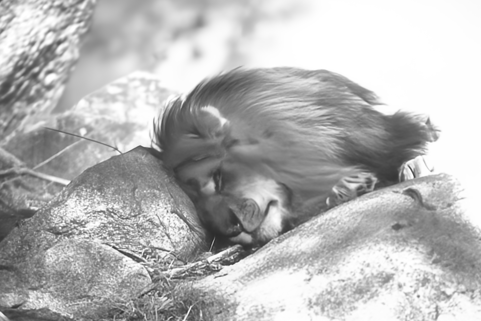}{27.66}\\[-3mm]
    
    \addlabel{TDV}
    \addcrop{\gls{bl-ift}}{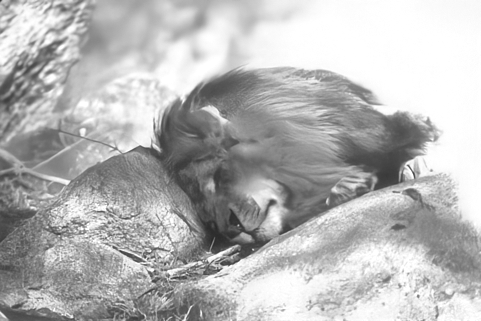}{27.59}
    \addcrop{\gls{bl-jfb}}{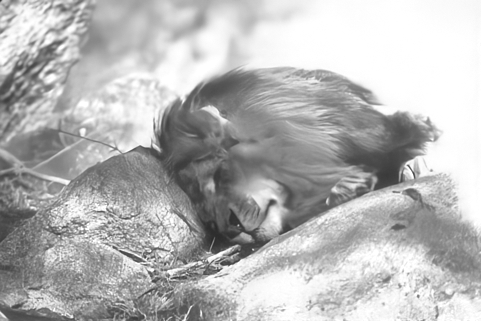}{27.59}
    \addcrop{\gls{ar}/\gls{lar}}{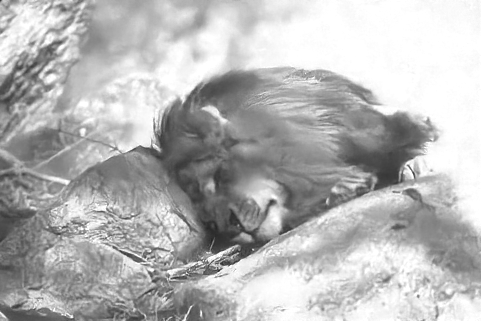}{27.10}
    \addcrop{\gls{sm}}{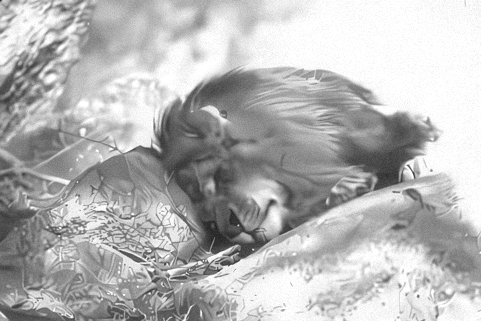}{26.69}
    \end{minipage}
    \endgroup
    \caption{Experiment 1: Comparison of training schemes on 'lion' image from BSD68.}
   \label{fig:denoise_bsd68:TDV}
\end{figure}
}

{
\setlength{\picheight}{4.4cm}
\setlength{\cropheight}{2.6cm}
\hypersetup{linkcolor=white}
\newcommand{\addfigure}[3]{%
    \begin{tikzpicture}%
    \node[anchor=south west,inner sep=0] (image) at (0,0) {\includegraphics[height=\picheight]{img/ct/exp2/#2.png}};%
    \begin{scope}[x={(image.south east)},y={(image.north west)}]%
    \node[text=white, anchor=north east] at (0.99,0.99) {\footnotesize \textbf{#3}};
    \node[text=white, anchor=south west] at (0.01,0.01) {\footnotesize \textbf{#1}};%
    \end{scope}%
    \end{tikzpicture}%
}
\newcommand{\addcrop}[3]{%
    \begin{tikzpicture}%
    \node[anchor=south west,inner sep=0] (image) at (0,0) {\includegraphics[height=\cropheight,trim={260 150 10 120}, clip]{img/ct/exp2/#2.png}};%
    \begin{scope}[x={(image.south east)},y={(image.north west)}]%
    \node[text=white, anchor=north east] at (0.99,0.99) {\footnotesize \textbf{#3}};
    \node[text=white, anchor=south west] at (0.01,0.01) {\footnotesize \textbf{#1\phantom{fg}}};%
    \end{scope}%
    \end{tikzpicture}%
}
\newcommand{\addlabel}[2]{%
\rotatebox{90}{\parbox[c]{#2}{\centering #1}\hspace*{-#2}}}
\begin{figure}
    \centering
    \begin{minipage}{\textwidth}
    \addlabel{Data}{\picheight}
    \addfigure{Ground Truth}{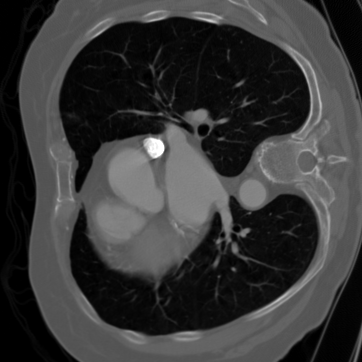}{$\infty$}
    \addfigure{\gls{fbp}}{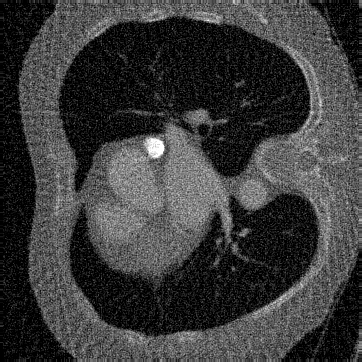}{24.15}\\[-3mm]

    \addlabel{Baseline}{\cropheight}
    \addcrop{Ground Truth}{ct4_gt_scale}{$\infty$}    
    \addcrop{\gls{fbp}}{ct4_fbp_scale}{24.15}
    \addcrop{\gls{fbp}+UNet}{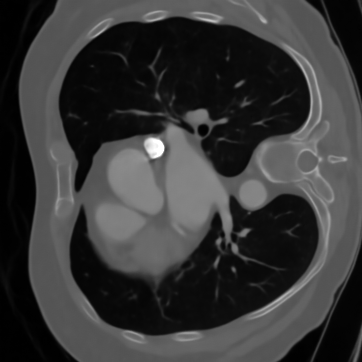}{39.14}
    \addcrop{\gls{lpd}}{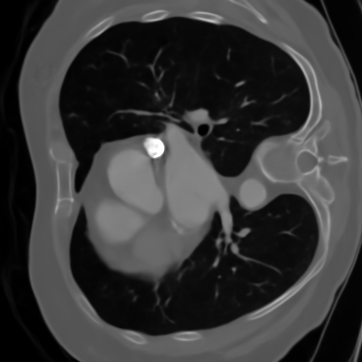}{39.78}\\[-3mm]

    \addlabel{Convex}{\cropheight}
    \addcrop{\gls{crr}}{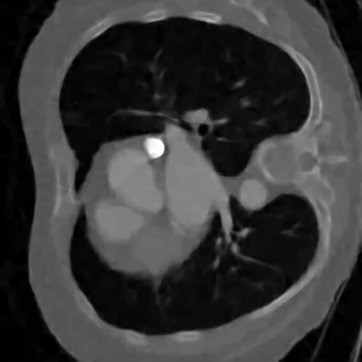}{37.19}
    \addcrop{\gls{icnn}}{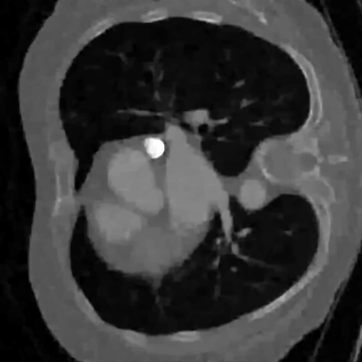}{36.82}\\[-3mm]

    \addlabel{\phantom{Nonconvex}}{\cropheight}    
    \addcrop{\gls{wcrr}}{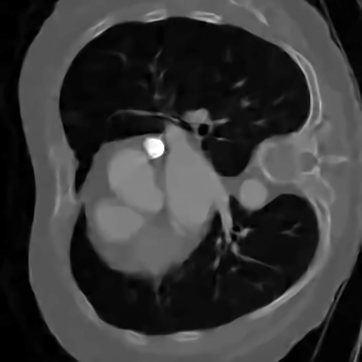}{38.11}
    \addcrop{\gls{idcnn}}{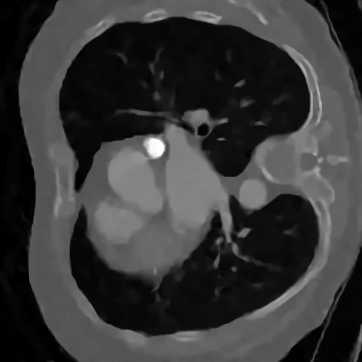}{37.52}     
    \addcrop{\gls{epll}}{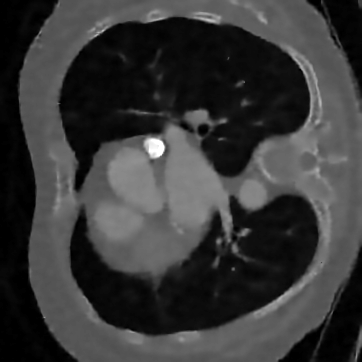}{36.85}
    \addcrop{\gls{patchnr}}{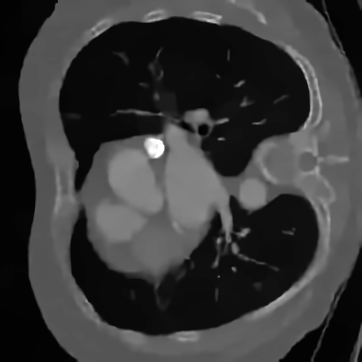}{37.37}\\[-3mm]    
    
    \addlabel{Nonconvex}{5cm}
    \addcrop{\gls{cnn}}{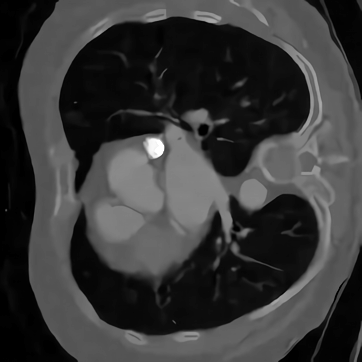}{37.22}
    \addcrop{\gls{tdv}}{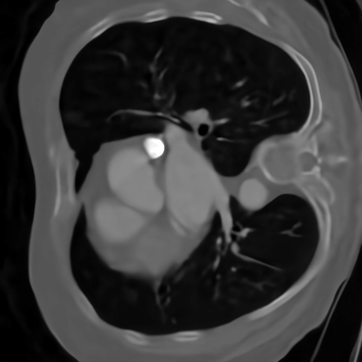}{38.90}
    \addcrop{\gls{lsr}}{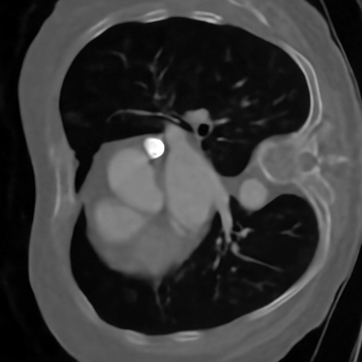}{38.84}
    \addcrop{\gls{lpn}}{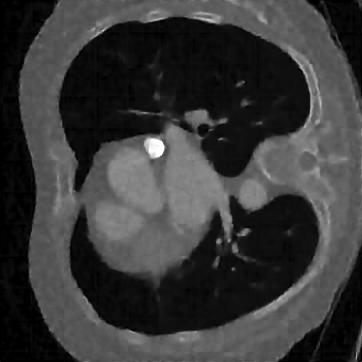}{35.94}
    \end{minipage}
    \caption{CT reconstructions for Experiment 2 on a test image from LoDoPaB-CT.}%
    \label{fig:ct_lodopab:transfer}
\end{figure}
}

{
\setlength{\picheight}{4.4cm}
\setlength{\cropheight}{2.6
cm}
\hypersetup{linkcolor=white}
\newcommand{\addfigure}[3]{%
    \begin{tikzpicture}%
    \node[anchor=south west,inner sep=0] (image) at (0,0) {\includegraphics[height=\picheight]{img/ct/exp3/#2.png}};%
    \begin{scope}[x={(image.south east)},y={(image.north west)}]%
    \node[text=white, anchor=north east] at (0.99,0.99) {\footnotesize \textbf{#3}};
    \node[text=white, anchor=south west] at (0.01,0.01) {\footnotesize \textbf{#1}};%
    \end{scope}%
    \end{tikzpicture}%
}
\newcommand{\addcrop}[3]{%
    \begin{tikzpicture}%
    \node[anchor=south west,inner sep=0] (image) at (0,0) {\includegraphics[height=\cropheight,trim={230 140 40 130}, clip]{img/ct/exp3/#2.png}};%
    \begin{scope}[x={(image.south east)},y={(image.north west)}]%
    \node[text=white, anchor=north east] at (0.99,0.99) {\footnotesize \textbf{#3}};
    \node[text=white, anchor=south west] at (0.01,0.01) {\footnotesize \textbf{#1\phantom{fg}}};%
    \end{scope}%
    \end{tikzpicture}%
}
\newcommand{\addlabel}[2]{%
\rotatebox{90}{\parbox[c]{#2}{\centering #1}\hspace*{-#2}}}
\begin{figure}
    \centering
    \begin{minipage}{\textwidth}
    \addlabel{Data}{\picheight}
    \addfigure{Ground Truth}{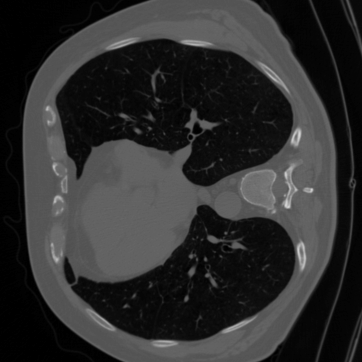}{$\infty$}
    \addfigure{\gls{fbp}}{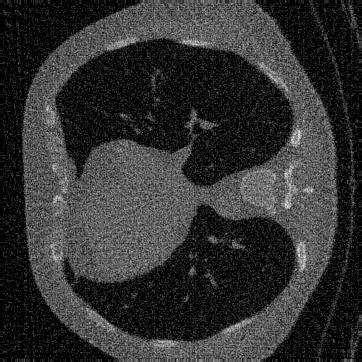}{19.22}\\[-3mm]

    \addlabel{Baseline}{\cropheight}
    \addcrop{Ground Truth}{ct9_gt_scale}{$\infty$}    
    \addcrop{\gls{fbp}}{ct9_fbp_scale}{19.22}
    \addcrop{\gls{fbp}+UNet}{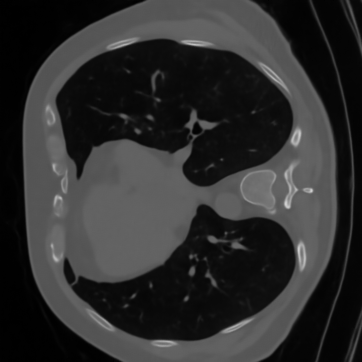}{35.15}
    \addcrop{\gls{lpd}}{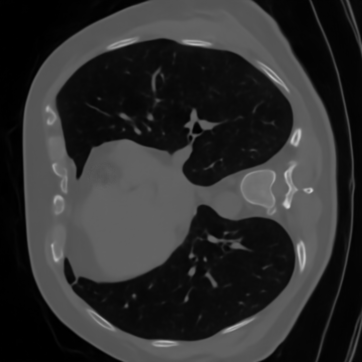}{35.47}\\[-3mm]

    \addlabel{Convex}{\cropheight}
    \addcrop{\gls{crr}}{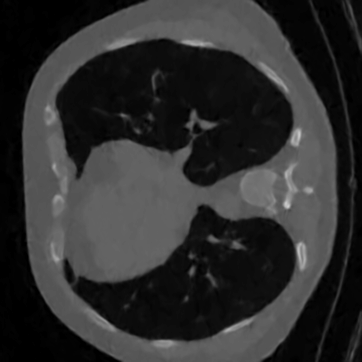}{32.84}
    \addcrop{\gls{icnn}}{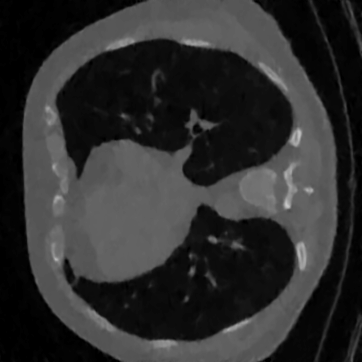}{32.52}\\[-3mm]

    \addlabel{\phantom{Nonconvex}}{\cropheight}    
    \addcrop{\gls{wcrr}}{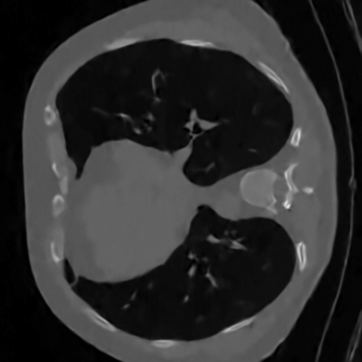}{34.06}
    \addcrop{\gls{idcnn}}{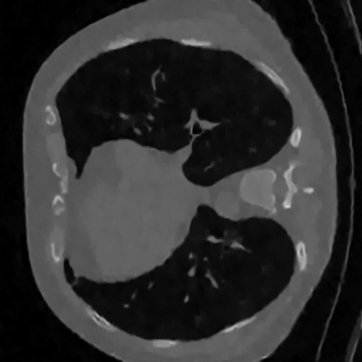}{33.39}     
    \addcrop{\gls{epll}}{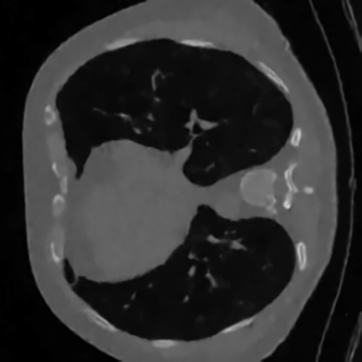}{33.58}
    \addcrop{\gls{patchnr}}{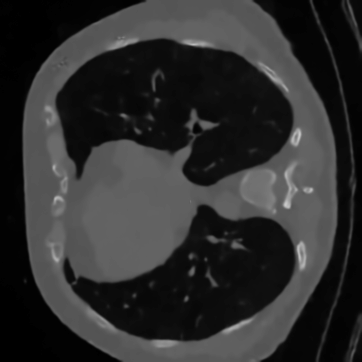}{33.78} \\[-3mm]    
    
    \addlabel{Nonconvex}{5cm}
    \addcrop{\gls{cnn}}{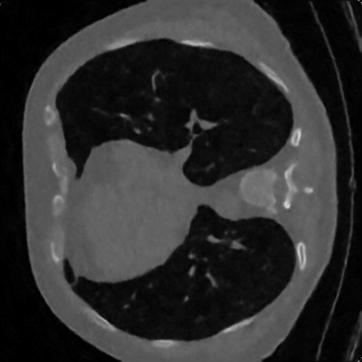}{32.60}
    \addcrop{\gls{tdv}}{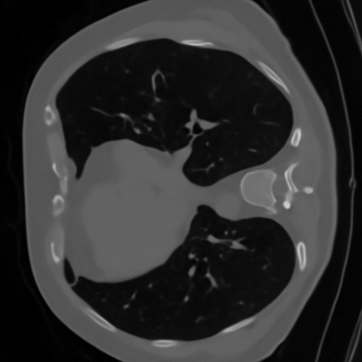}{35.26}
    \addcrop{\gls{lsr}}{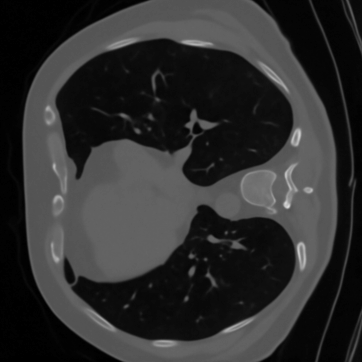}{35.53}
    \addcrop{\gls{lpn}}{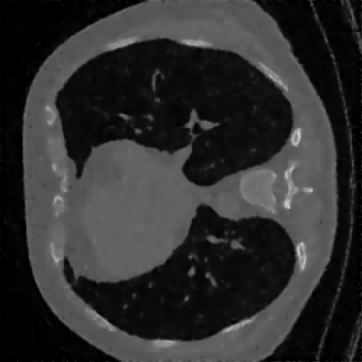}{32.63}
    \end{minipage}
    \caption{CT reconstructions for Experiment 3 on test image from LoDoPaB-CT.}%
   \label{fig:ct_lodopab}
\end{figure}
}

{
\setlength{\picheight}{4.4cm}
\setlength{\cropheight}{2.6cm}
\newcommand{\addfigure}[3]{%
    \begin{tikzpicture}%
    \node[anchor=south west,inner sep=0] (image) at (0,0) {\includegraphics[height=\picheight]{img/ct/lsr_comp/#2.png}};%
    \begin{scope}[x={(image.south east)},y={(image.north west)}]%
    \node[text=white, anchor=north east] at (0.99,0.99) {\footnotesize \textbf{#3}};
    \node[text=white, anchor=south west] at (0.01,0.01) {\footnotesize \textbf{#1}};%
    \end{scope}%
    \end{tikzpicture}%
}
\newcommand{\addcrop}[3]{%
    {%
    \hypersetup{linkcolor=white}%
    \begin{tikzpicture}%
    \node[anchor=south west,inner sep=0] (image) at (0,0) {\includegraphics[height=\cropheight,trim={230 140 40 130}, clip]{img/ct/lsr_comp/#2.png}};%
    \begin{scope}[x={(image.south east)},y={(image.north west)}]%
    \node[text=white, anchor=north east] at (0.99,0.99) {\footnotesize \textbf{#3}};
    \node[text=white, anchor=south west] at (0.01,0.01) {\footnotesize \textbf{#1\phantom{fg}}};%
    \end{scope}%
    \end{tikzpicture}%
    }%
}
\newcommand{\addlabel}[2]{%
\rotatebox{90}{\parbox[c]{#2}{\centering #1}\hspace*{-#2}}}
\begin{figure}
    \centering
    \begin{minipage}{\textwidth}
    \addlabel{Data}{\picheight}
    \addfigure{Ground Truth}{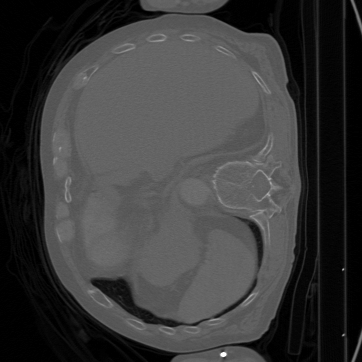}{$\infty$}
    \addfigure{Noisy}{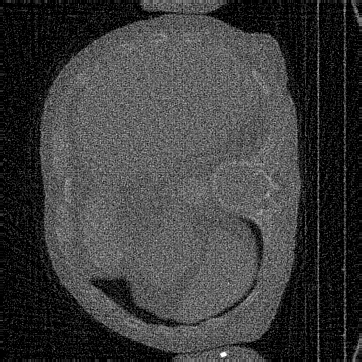}{24.07}\\[-3mm]
    
    \addlabel{Baseline}{\cropheight}
    \addcrop{Ground Truth}{ct8_gt_scale}{$\infty$}
    \addcrop{\gls{fbp}}{ct8_fbp_scale}{24.07}
    \addcrop{FBP+Unet}{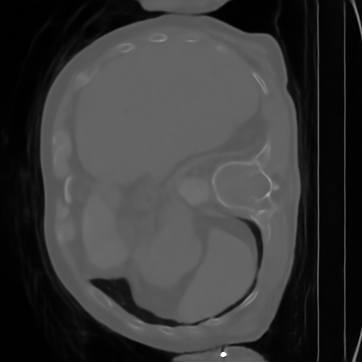}{36.38}
    \addcrop{\gls{lpd}}{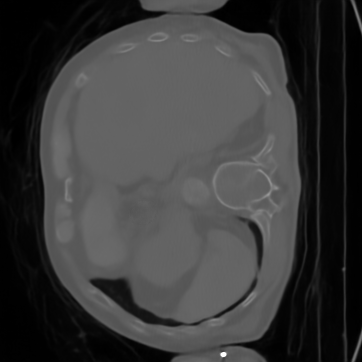}{38.71}
    \\[-3mm]
    
    \addlabel{LSR}{\cropheight}
    \addcrop{\gls{bl-jfb}+BSD}{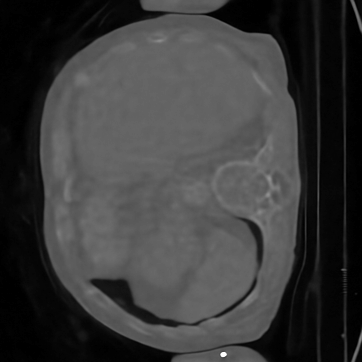}{38.40}
    \addcrop{\gls{bl-jfb}+CT}{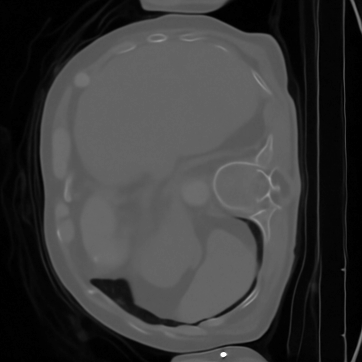}{38.89}
    \addcrop{\gls{nett}+BSD}{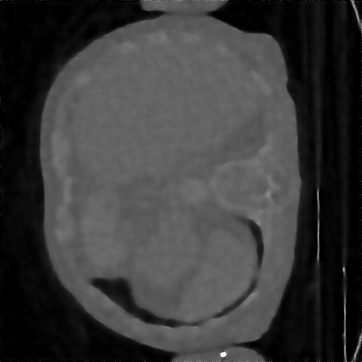}{34.91}
    \addcrop{\gls{nett}+CT}{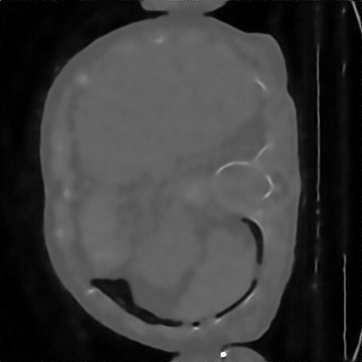}{35.50}
    \end{minipage}
    \caption{Comparison of training schemes to learn \gls{lsr} for \gls{ct} reconstruction.
    The test image is from LoDoPaB-CT, and the \enquote{+BSD} and \enquote{+CT} refer to Experiments~2 and 3, respectively.}
    \label{fig:ct_lodopab:comparetraining}
\end{figure}
}

\vspace{.2cm}

\begin{table}
\centering
\caption{Experiment 1: Training time in hours. \label{tab:bsd68_denoising_training_time}}
\setlength\tabcolsep{1.5pt}
{
\begin{tabular}{cl ccccc ccccc}
\toprule
& Architecture:
& \gls{crr} 
& \gls{icnn} 
& \gls{wcrr}  
& \gls{idcnn} 
& \gls{cnn} 
& \gls{tdv} 
& \gls{lsr} 
& \gls{epll} 
& \gls{patchnr} 
& \gls{lpn}\\ 
        
\midrule
\multirow{10}{1em}{\rotatebox{90}{Training Scheme}} 
& \gls{bl-ift} 
& \final{0.8}   
& \final{3.0}   
& \final{1.0}   
& \final{26.9}  
& \ILO          
& \final{11.6}  
& \final{41.2}  
& \WDW          
& \WDW          
& \WDW          
\\ 
& \gls{bl-jfb} 
& \final{0.5}   
& \final{0.8}   
& \final{0.7}   
& \final{2.7}   
& \final{7.3}   
& \final{7.1}   
& \final{13.3}  
& \WDW          
& \WDW          
& \WDW          
\\ 
& \gls{maid}
& \prelim{0.2}  
& \prelim{0.3}  
& \prelim{0.1}  
& \ILO          
& \ILO          
& \ILO          
& \ILO          
& \WDW          
& \WDW          
& \WDW          
\\ 
\cmidrule{2-12}
& \gls{ar}/\gls{lar}
& \prelim{0.2}  
& \final{0.5}   
& \prelim{0.2}  
& \final{4.5}   
& \final{2.7}   
& \prelim{3.1}  
& \ILO          
& \WDW          
& \WDW          
& \WDW          
\\
& \gls{nett}
& \WDW          
& \WDW          
& \WDW          
& \WDW          
& \WDW          
& \WDW          
& \prelim{11.0} 
& \WDW          
& \WDW          
& \WDW          
\\
\cmidrule{2-12}
& \gls{sm} 
& \final{0.1}   
& \final{0.3}   
& \final{0.1}   
& \final{0.9}   
& \final{0.3}   
& \final{3.6}   
& \final{9.2}   
& \WDW          
& \WDW          
& \WDW          
\\
& \gls{patchml}
& \WDW          
& \WDW          
& \WDW          
& \WDW          
& \WDW          
& \WDW          
& \WDW          
& \final{3.7}   
& \final{0.5}   
& \WDW          
\\
& \gls{pm} 
& \WDW          
& \WDW          
& \WDW          
& \WDW          
& \WDW          
& \WDW          
& \WDW          
& \WDW          
& \WDW          
& 15.2          
\\
\bottomrule
\end{tabular}
}
\vspace{.6cm}

\centering
\caption{Experiment 1: Reconstruction times per image in seconds. \label{tab:bsd68_denoising_time}}
\setlength\tabcolsep{1.5pt}
{
\begin{tabular}{cl ccccc ccccc}
\toprule
& Architecture:
& \gls{crr} 
& \gls{icnn} 
& \gls{wcrr}  
& \gls{idcnn} 
& \gls{cnn} 
& \gls{tdv} 
& \gls{lsr} 
& \gls{epll} 
& \gls{patchnr} 
& \gls{lpn}\\ 
        
\midrule
\multirow{10}{1em}{\rotatebox{90}{Training Scheme}} 
& \gls{bl-ift} 
& \final{0.80}   
& \final{0.19}   
& \final{0.46}   
& \final{0.19}   
& \ILO           
& \final{1.07}   
& \final{1.57}   
& \WDW           
& \WDW           
& \WDW           
\\ 
& \gls{bl-jfb} 
& \final{0.32}   
& \final{0.15}   
& \final{0.50}   
& \final{0.45}   
& \final{1.93}   
& \final{1.21}   
& \final{2.28}   
& \WDW           
& \WDW           
& \WDW           
\\ 
& \gls{maid}
& \prelim{0.67}  
& \prelim{0.12}  
& \prelim{0.32}  
& \ILO           
& \ILO           
& \ILO           
& \ILO           
& \WDW           
& \WDW           
& \WDW           
\\ 
\cmidrule{2-12}
& \gls{ar}/\gls{lar}
& \prelim{0.34}  
& \final{0.11}   
& \prelim{0.36}  
& \final{0.24}   
& \final{1.11}   
& \prelim{1.09}  
& \ILO           
& \WDW           
& \WDW           
& \WDW           
\\
& \gls{nett}
& \WDW           
& \WDW           
& \WDW           
& \WDW           
& \WDW           
& \WDW           
& 1.09           
& \WDW           
& \WDW           
& \WDW           
\\
\cmidrule{2-12}
& \gls{sm} 
& \final{0.26}   
& \final{0.09}   
& \final{0.28}   
& \final{0.31}   
& \final{1.17}   
& \final{1.71}   
& \final{4.34}   
& \WDW           
& \WDW           
& \WDW           
\\
& \gls{patchml}
& \WDW           
& \WDW           
& \WDW           
& \WDW           
& \WDW           
& \WDW           
& \WDW           
& 77.23          
& 5.37           
& \WDW           
\\
& \gls{pm} 
& \WDW           
& \WDW           
& \WDW           
& \WDW           
& \WDW           
& \WDW           
& \WDW           
& \WDW           
& \WDW           
& 0.27           
\\
\bottomrule
\end{tabular}
}
\end{table}

\section{Discussion}
\label{sec:discussion}
In Experiment 3, the larger architectures, such as \gls{tdv} and \gls{lsr} with bilevel training, yield the sharpest reconstructions and highest \gls{psnr} values.
Their performance is comparable with end-to-end reconstruction networks like \gls{fbp}+UNet or \gls{lpd}, which, however, cannot be an interpreted as variational reconstruction.
Moreover, the latter methods may introduce additional structures (hallucinations) in the reconstructions to resemble the training data more closely.
To a much lesser extent, this also occurs for \gls{tdv} and \gls{lsr}.
We did not detect such artifacts for the other regularizers used in our experimental comparison.
Thus, we can consider learned regularization as a robust and reliable method.

As discussed in detail below, another big advantage of learning regularizers is that they can be trained without using the operator $\forwardop$ and domain-specific data.
Hence, as a universal pretrained model, learned regularization can be a readily accessible tool in many applications, even when reliable ground truth data is not available.
Nevertheless, if computationally feasible, finetuning with task-specific data and the operator $\forwardop$ will improve the \gls{psnr} and visual reconstruction quality.

\paragraph*{Regularizer Architectures}
The discussed regularizers offer a trade-off in terms of theoretical guarantees, computational cost and expressivity.
Taking bilevel learning as an example, there is a large gap between convex architectures (like \gls{crr} and \gls{icnn}) and large nonconvex architectures (like \gls{tdv} and \gls{lsr}) in terms of visual reconstruction quality and \gls{psnr}.
Although slight relaxations of convexity (as in \gls{wcrr} and \gls{idcnn}) lead to improved performance, they cannot close the performance gap to large architectures.
Interestingly, the additional flexibility of \glspl{icnn} over the \gls{crr} seems to not improve the results in terms of \gls{psnr}. 
Our implementation of \gls{crr} requires fewer parameters, is easier to train and leads to better results.
An open question is whether other parameterizations of the \gls{icnn} can overcome this behavior.
Moreover, nonconvex extensions do not necessarily behave like their convex counterparts.
As an example, \gls{wcrr} aligns more closely with \gls{crr} than \gls{idcnn} with \gls{icnn}, especially regarding training time.

For denoising, we also compared with the learned DRUNet denoiser, whose performance is closely matched by \gls{tdv} and \gls{lsr} despite having an order of magnitude fewer parameters.
For the practically more relevant \gls{ct} reconstruction, both \gls{tdv} and \gls{lsr} are competitive with end-to-end reconstruction networks such as \gls{fbp}+UNet and \gls{lpd}.

\paragraph*{Training Methods}
Among the investigated training methods, \gls{bl-ift}/\gls{bl-jfb} consistently achieved the highest \gls{psnr} values.
However, it also requires the longest training times.
In this regard, \gls{ar} and \gls{sm} provide more efficient alternatives, both in terms of time and memory consumption.
Notably, for simple (convex) regularizers, these semi-supervised/unsupervised methods achieve reconstruction results that are comparable to those of \gls{bl-jfb}. 
This makes \gls{ar} a serious alternative for scaling these models to higher dimensions and larger datasets.
However, nearly optimal performance of \gls{ar} is already achieved by the \gls{wcrr} and more complex regularizers did not improve.
Thus, we regard \gls{bl-jfb}  as the gold standard when sufficient computational resources and supervised training data are available.
In particular, the results in Table~\ref{tab:numerics:domainshift} indicate that the learned models possess strong generalization capabilities.
As a consequence, one may train only for denoising and subsequently adjust the parameters $\alpha$ and $s$ (see Section \ref{subsec:experiments}), which is often substantially more efficient than training with the operator~$\forwardop$.

To further improve the computational efficiency, patch-based training of reconstruction networks is nowadays the standard for tasks where $\forwardop$ can be evaluated on patches.
In this case, the goal is not to approximate a patch distribution but to speed-up the training process.
Notably, the patches used in this context are typically much larger than those for methods discussed in Section \ref{Sec:PatchBased}.
For our experiments, we used patches of size between $25 \times 25$ and $80 \times 80$, depending on the field-of-view of the regularizer.
While we found patches to be beneficial for all presented methods, they were particularly important for \gls{ar} training in the setting of \gls{ct} reconstruction.
In summary, we advise to always train on patches if possible, both from a stability and efficiency perspective.

We conclude with a remark on \gls{nett}.
There, both the training/validation losses decrease consistently throughout training, but the reconstruction quality often deteriorates beyond some point.
This behavior is even more severe than for \gls{ar}, and a reason for this could be the lack of a regularizing term in the loss \eqref{eq:NETTloss} compared to \eqref{eq:AR_loss}.
The latter helps to prevent overfitting.

\paragraph*{Different Variants of Bilevel Learning}
Across our experiments, all bilevel methods have led to comparable reconstruction performance. 
In practice, the \gls{ift} mode is only feasible if the Hessian-vector products can be computed efficiently, or in strongly convex settings where linear solvers converge in a few steps.
Otherwise, it is computationally infeasible.
Since the computationally much more efficient \gls{jfb} mode consistently matched or even outperformed the \gls{ift} mode, we recommend it as the default choice.

For the \gls{ift} and \gls{jfb} modes, we found it beneficial to employ \gls{sm} pretraining together with Hessian regularization, in line with prior observations \cite{ZouLiuWoh2023}.
The latter also reduces the number of steps required by the \gls{nmapg} during evaluation, a property that is not promoted by the plain bilevel objective \eqref{eq:bilevel-prob}.
After incorporating these elements, the results were consistent across multiple training runs.
This contrasts \gls{ar} and \gls{nett}, both of which exhibit high variance between runs.
A source for this behavior could be the relatively small datasets, where overfitting a classifier is likely.

The \gls{jfb} and \gls{ift} mode both assume convergence of the involved iterative solvers, which requires hand-tuning the corresponding accuracies and step sizes.
\Gls{maid} provides a method to choose them adaptively, which can speed up the optimization and guarantees convergence.
However, currently it does not support stochastic optimization in the upper-level solver and the theoretical analysis requires the lower-level problem to be convex.
Moreover, it relies on the computationally more expensive \gls{ift} mode.
Therefore, it is only applicable for small (weakly) convex architectures for which the numerical results from the previous sections show similar performance as for \gls{jfb} and \gls{ift}.
Addressing these shortcomings remains subject of future work.

\paragraph*{Solving the Variational Problem}

To keep the comparison manageable, we only used \gls{nmapg} to solve the variational problem \eqref{eq:VarProb}, which worked well for all the tested regularizers. 
From an application perspective, a fast minimization of \eqref{eq:VarProb} is key to scaling the approach and it plays an important role during the bilevel training approaches.
In this chapter, however, we placed the focus on reconstruction quality, and faster convergence was not encouraged during training.
A comparison with other algorithms would be interesting but is beyond the scope of this chapter.

Once the regularizer $R$ is fixed, one can investigate properties of the variational problem~\eqref{eq:VarProb}, and in particular, the properties of the resulting variational solution itself.
For the \gls{nett} \cite{LiSch2020} and \gls{ar} \cite{lunz2018adversarial,shumaylov2024weakly} frameworks, it has been shown that plugging the learned regularizer into \eqref{eq:VarProb} yields a well-posed regularization method: solutions exist, the data-to-reconstruction map is continuous, and suitable parameter choice rules lead to convergence for vanishing noise.
Stronger continuity results in the measurement domain can be obtained for convex regularizers \cite[Prop.\ 3.2]{GouNeuBoh2022}.
A related result involving spatially varying $\Lambda$ (see Section \ref{sec:FoE}) has been derived in \cite[Prop.\ 6]{KofAltBa2023} for conditional \gls{tv} regularization.
Several generalizations and an extension to uncertainty in the data $\vec y $ itself can be found in \cite{NeuAlt2024}.
For a recent overview, we refer to \cite{MukHauOek2023}.

\paragraph*{Limitations}

This chapter provides a comparison of regularizer architectures and training schemes.
To this end, we unified the training and evaluation setting as much as possible.
Due to time constraints, certain aspects are not investigated systematically.
In particular, we did not examine how the reconstruction performance depends on the dataset size.
Furthermore, in many practical applications the operator $\forwardop$ is subject to modeling errors and the Gaussian noise assumption may not hold; robustness of the learned regularizers to such settings is not addressed.
For bilevel learning, alternative upper-level loss functions have been proposed, including \gls{mse}, $L_1$, \gls{pm}, LPIPS \cite{zhang2018unreasonable}, and \gls{tv} \cite{de2017bilevel}.
These might affect the reconstruction performance noticeably.
Likewise, methods for quantifying uncertainty in reconstructions, though highly relevant, were not considered.

We also note that not all entries are filled in Tables~\ref{tab:bsd68_denoising}, \ref{tab:numerics:domainshift}, and \ref{tab:numerics:CT}.
Some regularizers (e.g., \gls{epll}, \gls{patchnr}, \gls{lpn}) can only be trained in specific ways.
For \gls{lsr}, training occasionally exhibited instabilities that prevented further comparisons.
The training algorithm \gls{maid} requires convexity of the variational problem \eqref{eq:VarProb}, which is only fulfilled for convex or weakly-convex regularizers.
Generalizations remain outside the scope of this chapter.

\paragraph*{Extensions}

In this chapter we focused on the comparison of existing learned regularizers and training methods.
Interesting extensions of the current methods are the inclusion of the noise level as input of the regularizer, or properties of the operator $\forwardop$ such as source conditions \cite{MukSch2021}.
It would also be interesting to investigate if there is a benefit of training a regularizer on several inverse problems at once, i.e., can we learn a foundational model for general inverse problems. 
Further, a systematic analysis and evaluation of learned regularizers with respect to their theoretical properties would be highly valuable.
More broadly, a fundamental open question is what theoretical properties are most desirable for learned regularizers, and whether the current -- primarily functional analytic -- viewpoint is the most appropriate for regularizers trained on finite-dimensional datasets.

\section{Conclusions}
\label{sec:conclusions}
In this chapter, we examined the learning of variational regularization functionals for inverse problems.
We outlined the core ideas of each method, compared their strengths and limitations, and showed that most outperform traditional hand-crafted regularizers while retaining interpretability and stability.
In practice, one has to balance between theoretical properties (e.g., weak convexity), computational cost, requirements on the training dataset, and reconstruction quality.
Our study highlights these trade-offs, and demonstrates the performance gains enabled by more flexible architectures and increased compute resources.
To ensure reproducibility, we provided key implementation details, and released the training and evaluation code online.

Looking ahead, several challenges and opportunities remain open. On the theoretical side, understanding of generalization and stability will be essential. On the practical side, more efficient training and optimization strategies are needed to enable large-scale deployment. Further, incorporating uncertainty quantification and exploring task-specific models may open promising avenues for broader applicability.

\section*{Acknowledgments}

MJE acknowledges support from the EPSRC (EP/T026693/1; EP/Y037286/1).
ZK acknowledges support from the EPSRC (EP/X010740/1).
AD, CBS, HSW and MJE acknowledge support from the EPSRC (EP/V026259/1).
JH acknowledges funding from the DFG (530824055).
EK, SN and GSW acknowledge support from the DFG (SPP2298 - 543939932). 
EK acknowledges funding from the FWF (10.55776/COE12).
SD acknowledges funding from the ERC (101020573 FunLearn).
CBS acknowledges support from the Royal Society Wolfson Fellowship, the EPSRC (EP/V029428/1; ProbAI hub EP/Y028783/1), and the Wellcome Innovator Awards (215733/Z/19/Z; 221633/Z/20/Z).
MJE and CBS acknowledge support from the EU Horizon 2020 research and innovation programme under the Marie Skodowska-Curie grant agreement REMODEL.
JS and ZF acknowledge funding from NIH Grant P41EB031771.

\printnoidxglossaries

\bibliographystyle{abbrv}
\bibliography{references}
\end{document}